\documentclass[sigconf, nonacm]{acmart}

\AtBeginDocument{%
  }

\usepackage{graphicx}   % 支持插图
\usepackage{subcaption}  % 推荐这个比 subfigure 更新
\usepackage{float}
\usepackage{multirow}
\usepackage{algorithm}
\usepackage{algpseudocode}
\begin{document}

\title{GeoGNN: Quantifying and Mitigating Semantic Drift in Text-Attributed Graphs}

% ---- Author Information ----
% 让 hyperref 生成书签时忽略格式命令（防止警告）
\pdfstringdefDisableCommands{%
  \def\textsuperscript#1{#1}%
  \def\nolinkurl#1{#1}%
}
\author{Liangwei Yang\textsuperscript{1*}, Jing Ma\textsuperscript{2}, Jianguo Zhang\textsuperscript{1}, Zhiwei Liu\textsuperscript{1},
  Jielin Qiu\textsuperscript{1}, \\
  Shirley Kokane\textsuperscript{1},  Shiyu Wang\textsuperscript{1},
  Haolin Chen\textsuperscript{1}, Rithesh Murthy\textsuperscript{1}, Ming Zhu\textsuperscript{1}, \\
  Huan Wang\textsuperscript{1}, Weiran Yao\textsuperscript{1}, Caiming Xiong\textsuperscript{1}, Shelby Heinecke\textsuperscript{1}\\[3pt]
  \textsuperscript{1}Salesforce AI Research, Palo Alto, CA, USA\\
  \textsuperscript{2}Independent Researcher, Palo Alto, CA, USA}

% —— 直接覆盖 acmart 的内部作者宏（放在 \maketitle 之前！）——
% \makeatletter
% \gdef\@authors{%
%   Liangwei Yang\textsuperscript{1*}, Jing Ma\textsuperscript{2}, Jianguo Zhang\textsuperscript{1}, Zhiwei Liu\textsuperscript{1},
%   Jielin Qiu\textsuperscript{1}, Shirley Kokane\textsuperscript{1}, Shiyu Wang\textsuperscript{1},
%   Haolin Chen\textsuperscript{1}, Rithesh Murthy\textsuperscript{1}, Ming Zhu\textsuperscript{1},
%   Huan Wang\textsuperscript{1}, Weiran Yao\textsuperscript{1}, Caiming Xiong\textsuperscript{1}, Shelby Heinecke\textsuperscript{1}\\[3pt]
%   \textsuperscript{1}Salesforce AI Research, Palo Alto, CA, USA\\
%   \textsuperscript{2}Independent Researcher, USA
% }
% \makeatother

\renewcommand{\shortauthors}{Liangwei et al.}

\begin{abstract}
Graph neural networks (GNNs) on text-attributed graphs (TAGs) typically encode node texts using pretrained language models (PLMs) and propagate these embeddings through linear neighborhood aggregation. However, the representation spaces of modern PLMs are highly non-linear and geometrically structured, where textual embeddings reside on curved semantic manifolds rather than flat Euclidean spaces. Linear aggregation on such manifolds inevitably distorts geometry and causes \emph{semantic drift}—a phenomenon where aggregated representations deviate from the intrinsic manifold, losing semantic fidelity and expressive power. To quantitatively investigate this problem, this work introduces a local PCA-based metric that measures the degree of semantic drift and provides the first quantitative framework to analyze how different aggregation mechanisms affect manifold structure. Building upon these insights, we propose \textbf{Geodesic Aggregation}, a manifold-aware mechanism that aggregates neighbor information \emph{along geodesics} via log–exp mappings on the unit sphere, ensuring that representations remain faithful to the semantic manifold during message passing. We further develop \textbf{GeoGNN}, a practical instantiation that integrates spherical attention with manifold interpolation. Extensive experiments across four benchmark datasets and multiple text encoders show that GeoGNN substantially mitigates semantic drift and consistently outperforms strong baselines, establishing the importance of manifold-aware aggregation in text-attributed graph learning.
\end{abstract}

\begin{CCSXML}
<ccs2012>
   <concept>
       <concept_id>10010405.10010497</concept_id>
       <concept_desc>Applied computing~Document management and text processing</concept_desc>
       <concept_significance>500</concept_significance>
       </concept>
   <concept>
       <concept_id>10002951.10003260</concept_id>
       <concept_desc>Information systems~World Wide Web</concept_desc>
       <concept_significance>500</concept_significance>
       </concept>
   <concept>
       <concept_id>10010147.10010257</concept_id>
       <concept_desc>Computing methodologies~Machine learning</concept_desc>
       <concept_significance>500</concept_significance>
       </concept>
 </ccs2012>
\end{CCSXML}

\ccsdesc[500]{Applied computing~Document management and text processing}
\ccsdesc[500]{Information systems~World Wide Web}
\ccsdesc[500]{Computing methodologies~Machine learning}

\keywords{Semantic Drift, Graph Neural Network, Text-attributed Graphs}



\maketitle

\section{Introduction}

Graph neural networks (GNNs) have become a cornerstone of representation learning for graph-structured data~\cite{hamilton2017inductive,yang2022large,yang2023dgrec,liu2024knowledge}. In text-attributed graphs (TAGs), where each node is associated with rich textual content, recent studies typically encode node texts using pretrained language models (PLMs) such as BERT~\cite{devlin2019bert}, RoBERTa~\cite{liu2019roberta}, or MiniLM~\cite{wang2020minilm}, and then propagate these embeddings through neighborhood aggregation~\cite{he2022gnntext}. 
Despite their empirical success, most GNNs still rely on \emph{linear aggregation} (e.g., mean, Laplacian, or attention-based averaging) to combine neighbor features—an assumption that has been largely taken for granted.

\begin{figure}[htbp]  % t=top, h=here, b=bottom, p=page of floats
    \centering
    \includegraphics[width=0.9\linewidth]{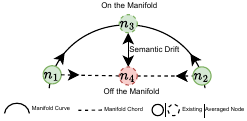}
    \caption{Illustration of Semantic Drift: linear averaging ($n_4$) deviates from the semantic manifold, while geodesic averaging ($n_3$) stays on the manifold.}
    \label{pics:drift}
\end{figure}

However, the representation space produced by modern PLMs is highly \emph{non-linear and geometrically structured}~\cite{ethayarajh2019contextual, timkey2021all}. 
Empirical studies reveal that textual embeddings lie on curved semantic manifolds rather than flat Euclidean spaces. 
When GNNs perform linear neighborhood aggregation on such manifolds, they implicitly assume local linearity and thus risk distorting the manifold geometry. 
We term this phenomenon \textbf{Semantic Drift}: the aggregated representation deviates from the intrinsic semantic manifold, losing fidelity to the original feature space and reducing expressive power. 
As illustrated in Figure~\ref{pics:drift}, linear averaging pulls a node representation (e.g., $n_4$) off the curved manifold toward the Euclidean mean of its neighbors, while our geodesic aggregation follows the manifold geometry (e.g., $n_3$), preserving semantic consistency. 
This drift accumulates across layers, gradually flattening the geometry of text embeddings and leading to degraded downstream performance.

As pretrained language models continue to evolve, their latent semantic manifolds become increasingly rich and structured. 
Modern encoders already capture semantically meaningful vector representations, effectively embedding linguistic structures into the geometry of latent space~\cite{reimers2019sentence}. 
Consequently, the aggregation step—traditionally viewed as a simple neighborhood smoother—now plays a more delicate role: if it fails to respect the geometry of the encoder’s latent manifold, the resulting representations may lose semantic information that the encoder has worked to preserve.

To systematically investigate this phenomenon, we introduce a local PCA-based metric that quantitatively measures \textbf{semantic drift}—the degree to which aggregated embeddings deviate from their underlying semantic manifolds. 
This metric allows us to empirically analyze how different aggregation mechanisms distort manifold geometry, offering the first quantitative framework for studying semantic drift in text-attributed graphs.
Building on these insights, we propose \textbf{Geodesic Aggregation}, a manifold-aware mechanism that aggregates neighbor information \emph{along geodesics} rather than through linear averaging. 
By leveraging log–exp mappings on the unit sphere, our method ensures that updated representations remain on the semantic manifold throughout message passing. 
Based on this principle, we develop \textbf{GeoGNN}, a practical instantiation that integrates spherical attention with geodesic interpolation on the manifold.
Extensive experiments across four benchmark datasets and multiple text encoders show that GeoGNN effectively mitigates semantic drift and consistently outperforms strong baselines, underscoring the importance of manifold-aware aggregation in text-attributed graph learning.
In summary, our contributions are threefold:
\begin{itemize}
    \item We identify and formalize the problem of \textbf{semantic drift} in text-attributed graph neural networks, arising from linear aggregation on non-Euclidean manifolds.
    \item We propose \textbf{Geodesic Aggregation}, a novel mechanism that performs neighbor aggregation along geodesics via log–exp mappings, ensuring manifold fidelity.
    \item We introduce \textbf{GeoGNN}, an instantiation of our approach, and a PCA-based metric to quantify semantic drift. Extensive experiments validate the effectiveness and generality of our framework.
\end{itemize}

\section{Semantic Drift in Text-Attributed Graphs}
\subsection{Text-Attributed Graphs and Learning Tasks}

We consider a \textbf{text-attributed graph (TAG)} 
$\mathcal{G} = (\mathcal{V}, \mathcal{E}, \mathbf{X})$, 
where $\mathcal{V}$ is the node set, $\mathcal{E}$ the edge set, 
and each node $v_i \in \mathcal{V}$ has a textual description encoded by a pretrained language model (PLM) into an embedding $\mathbf{x}_i \in \mathbb{R}^d$. 
These embeddings serve as initial node features for graph neural networks (GNNs).

\noindent\textbf{Message Passing on TAGs.}
Let $\mathbf{h}_i^{(0)}=\mathbf{x}_i$. 
At layer $l$, denote the neighbor set by $\mathcal{N}(i)$. 
A generic GNN layer first aggregates neighbor features to obtain an intermediate representation:
\begin{equation}
\bar{\mathbf{h}}_i^{(l+1)} 
= \sum_{j \in \mathcal{N}(i)} w_{ij}\, \mathbf{h}_j^{(l)},
\label{eq:linear_agg}
\end{equation}
where $w_{ij}$ are pre-defined or learned normalized weights (e.g., mean/Laplacian/attention). 
Then a learnable update function produces the next-layer embedding:
\begin{equation}
\mathbf{h}_i^{(l+1)} 
= \phi\!\big(\mathbf{h}_i^{(l)},\, \bar{\mathbf{h}}_i^{(l+1)}\big),
\label{eq:update}
\end{equation}
where $\phi(\cdot)$ can be an MLP with nonlinearity. 
After $L$ layers we obtain $\mathbf{h}_i^{(L)}$ for downstream tasks.

\noindent\textbf{Node Classification.}
Given a labeled subset $\mathcal{V}_L \subset \mathcal{V}$ with labels $y_i \in \{1,\dots,C\}$, 
predictions are made from the embeddings via:
\begin{equation}
\hat{\mathbf{y}}_i = \operatorname{softmax}\!\big(\mathbf{W}_o \mathbf{h}_i^{(L)}\big),
\label{eq:classification}
\end{equation}
and the model is trained with cross-entropy on $\mathcal{V}_L$.

\noindent\textbf{Link Prediction.}
Given a pair $(u,v)$, we score the likelihood that an edge exists between them:
\begin{equation}
p(u,v) = \sigma\!\big(\operatorname{sim}(\mathbf{h}_u^{(L)}, \mathbf{h}_v^{(L)})\big),
\label{eq:link_prediction}
\end{equation}
where $\operatorname{sim}(\cdot,\cdot)$ is typically a dot product or cosine similarity, 
and $\sigma$ is the logistic function. 
Training uses positive (observed) and negative (sampled) node pairs.

\noindent\textbf{Aggregation and Its Limitation.}
Most existing TAG models adopt linear neighborhood aggregation---%
mean, Laplacian, or attention-based averaging---%
implicitly assuming the embedding space $\mathbb{R}^d$ is Euclidean and locally flat.
However, recent studies~\cite{ethayarajh2019contextual, timkey2021all} 
show that embeddings from PLMs reside on highly curved \emph{semantic manifolds}.
Applying linear aggregation on such manifolds violates this geometric structure, 
causing distortions in node representations and harming both node classification and link prediction performance.

\subsection{Defining Semantic Drift}

Building upon the standard linear aggregation in Eq.~\ref{eq:linear_agg}, 
we consider the geometric implications of message passing 
when the feature space is a curved manifold rather than a flat Euclidean space.
Let $\mathbf{h}_i^{(l)} \in \mathbb{R}^d$ denote the representation of node $v_i$
on a semantic manifold $\mathcal{M} \subset \mathbb{R}^d$ 
induced by the pretrained language model (PLM). 
The aggregated feature $\bar{\mathbf{h}}_i^{(l+1)}$ computed in Eq.~\ref{eq:linear_agg} 
is a Euclidean mean of neighbor features and thus generally lies \emph{off the manifold} $\mathcal{M}$.

We define the deviation between this Euclidean aggregation 
and its closest point on the manifold as \textbf{Semantic Drift}.

\begin{definition}[Semantic Drift]
Given a manifold $\mathcal{M}$ embedded in $\mathbb{R}^d$, 
the semantic drift of node $v_i$ after aggregation is
\begin{equation}
D_{\text{drift}}(i) 
= \operatorname{dist}_\mathcal{M}\!\left(
  \bar{\mathbf{h}}_i^{(l+1)},\,
  \Pi_\mathcal{M}\!\big(\bar{\mathbf{h}}_i^{(l+1)}\big)
\right),
\label{eq:semantic_drift}
\end{equation}
where $\Pi_\mathcal{M}(\cdot)$ denotes the projection onto $\mathcal{M}$, 
and $\operatorname{dist}_\mathcal{M}$ is the geodesic distance on $\mathcal{M}$.
\end{definition}

Intuitively, semantic drift quantifies how far the aggregated embedding 
has moved away from the intrinsic semantic manifold defined by the PLM. 
As illustrated in Figure~\ref{pics:drift}, 
linear averaging (e.g., node $n_4$) pulls the representation 
toward the Euclidean mean of its neighbors, moving it off the manifold,
while geodesic aggregation (e.g., node $n_3$) follows the curved geometry, 
preserving semantic consistency and manifold fidelity.
This drift accumulates over layers, gradually flattening the semantic geometry of textual embeddings.
In the next subsection, we propose a quantitative metric to measure and analyze this phenomenon in practice.

\subsection{Quantifying Semantic Drift}

\begin{figure*}[t]
  \centering
  % --- 四张子图 ---
  \begin{subfigure}{0.24\linewidth}
    \centering
    \includegraphics[width=\linewidth]{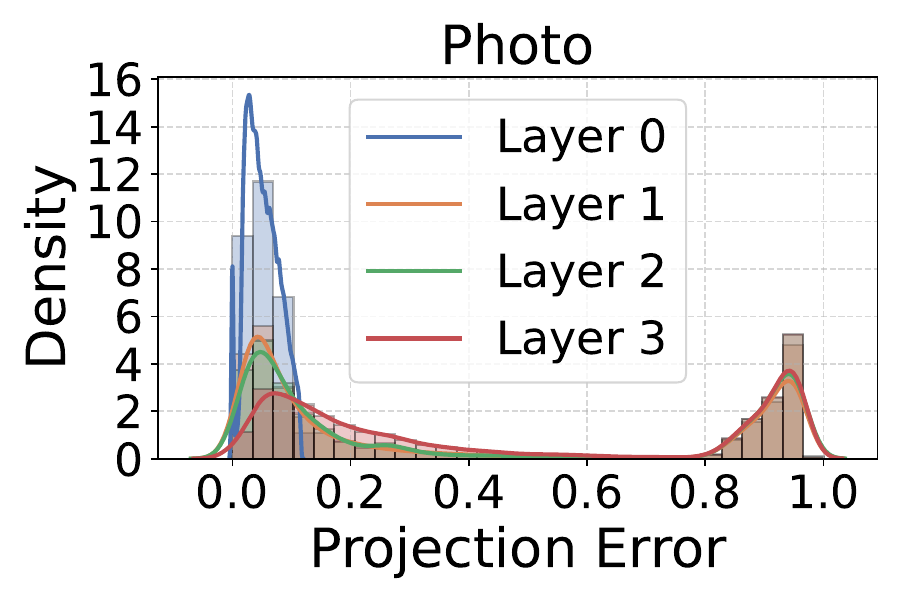}
    \caption{Mean Smoothing}
  \end{subfigure}
  \begin{subfigure}{0.24\linewidth}
    \centering
    \includegraphics[width=\linewidth]{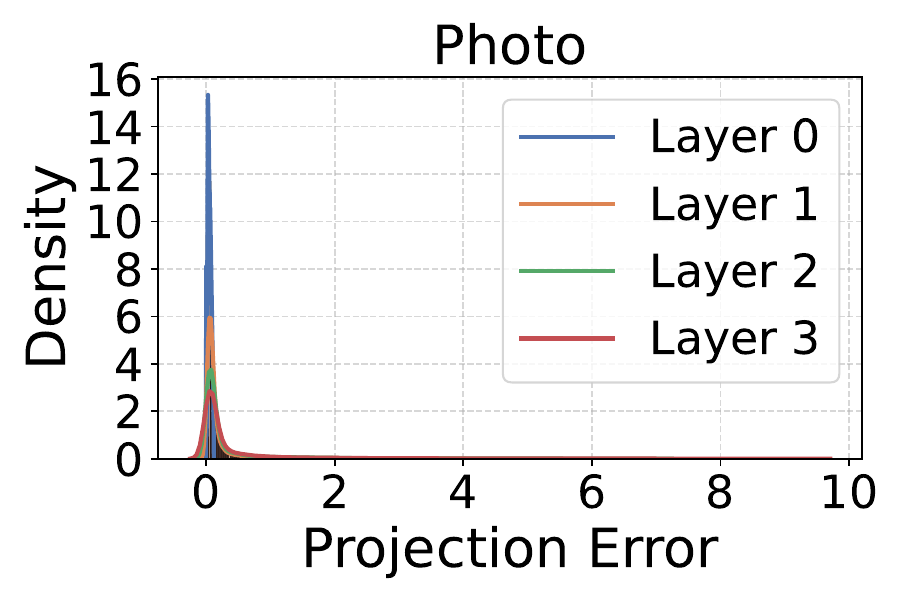}
    \caption{Laplacian Smoothing}
  \end{subfigure}
  \begin{subfigure}{0.24\linewidth}
    \centering
    \includegraphics[width=\linewidth]{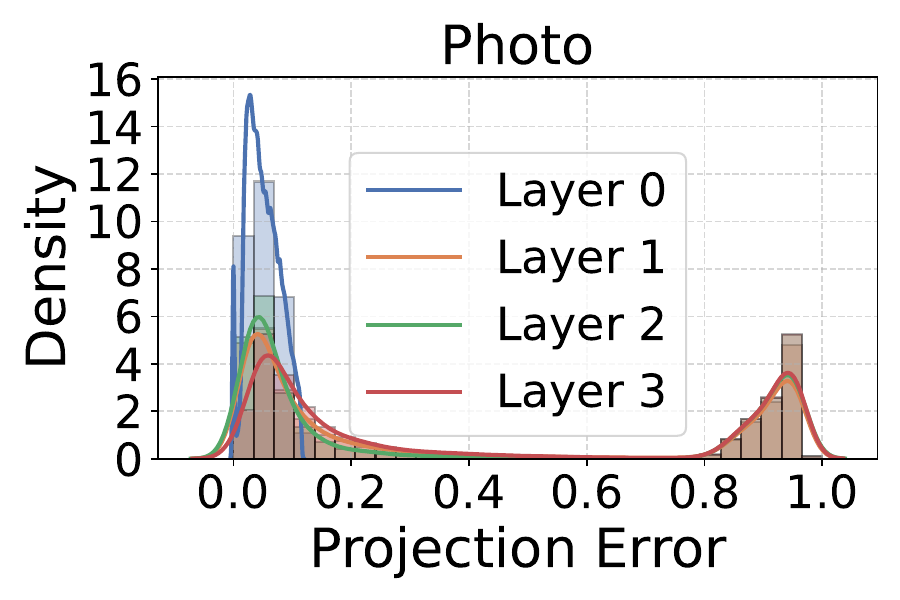}
    \caption{Attention Smoothing}
  \end{subfigure}
  \begin{subfigure}{0.24\linewidth}
    \centering
    \includegraphics[width=\linewidth]{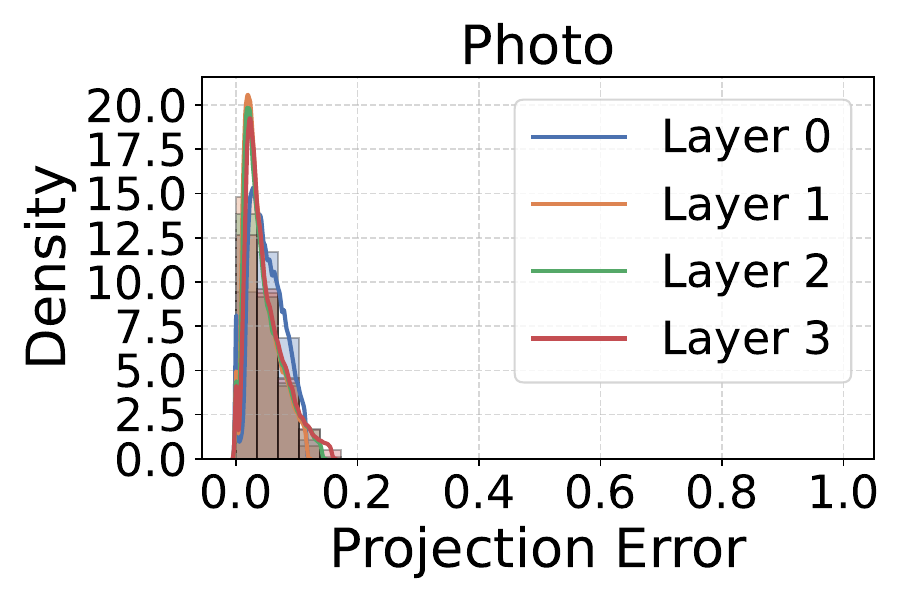}
    \caption{Geodesic Smoothing}
  \end{subfigure}

  % --- 四张子图 ---
  \begin{subfigure}{0.24\linewidth}
    \centering
    \includegraphics[width=\linewidth]{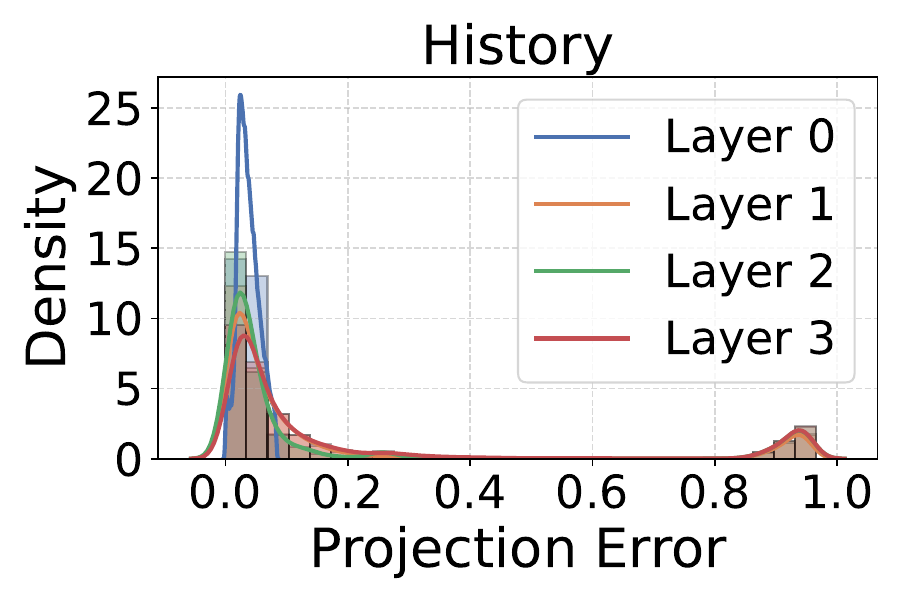}
    \caption{Mean Smoothing}
  \end{subfigure}
  \begin{subfigure}{0.24\linewidth}
    \centering
    \includegraphics[width=\linewidth]{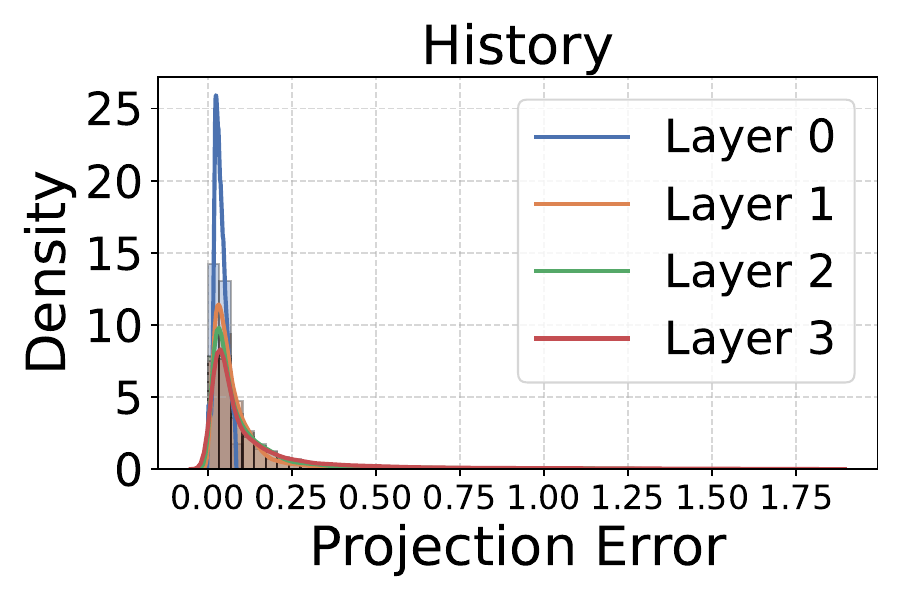}
    \caption{Laplacian Smoothing}
  \end{subfigure}
  \begin{subfigure}{0.24\linewidth}
    \centering
    \includegraphics[width=\linewidth]{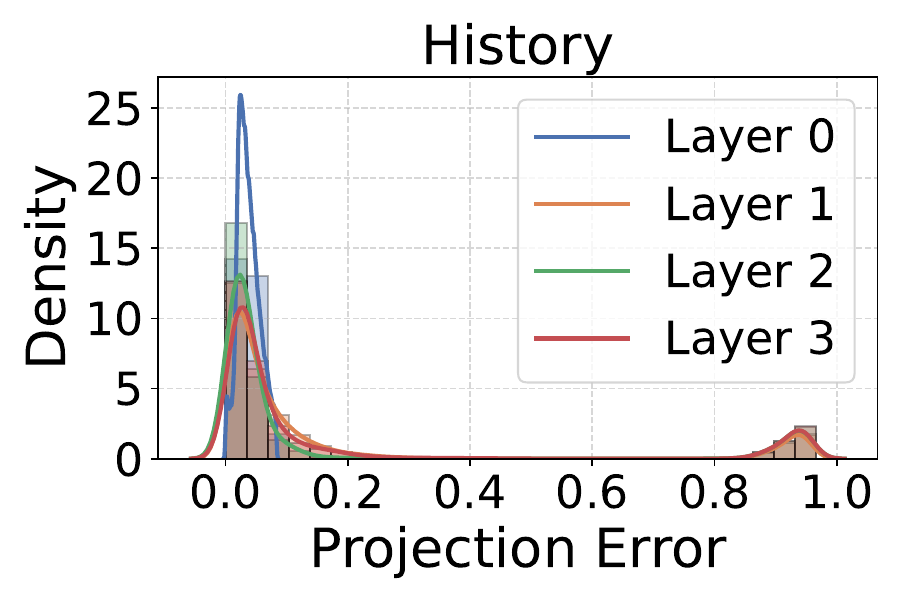}
    \caption{Attention Smoothing}
  \end{subfigure}
  \begin{subfigure}{0.24\linewidth}
    \centering
    \includegraphics[width=\linewidth]{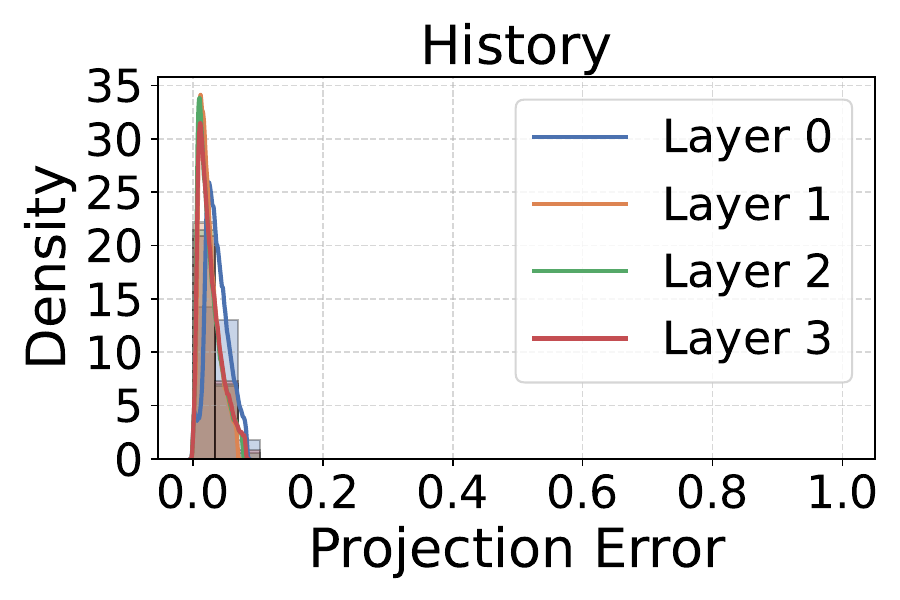}
    \caption{Geodesic Smoothing}
  \end{subfigure}
  \caption{Comparison of four aggregators (Mean, Laplacian, Attention, Geodesic) on Photo and History dataset. Our Geodesic Aggregator preserves the manifold structure and mitigates semantic drift.}
  \label{fig:agg_comparison}
\end{figure*}

\begin{figure}[H]
  \centering
  \includegraphics[width=0.485\linewidth]{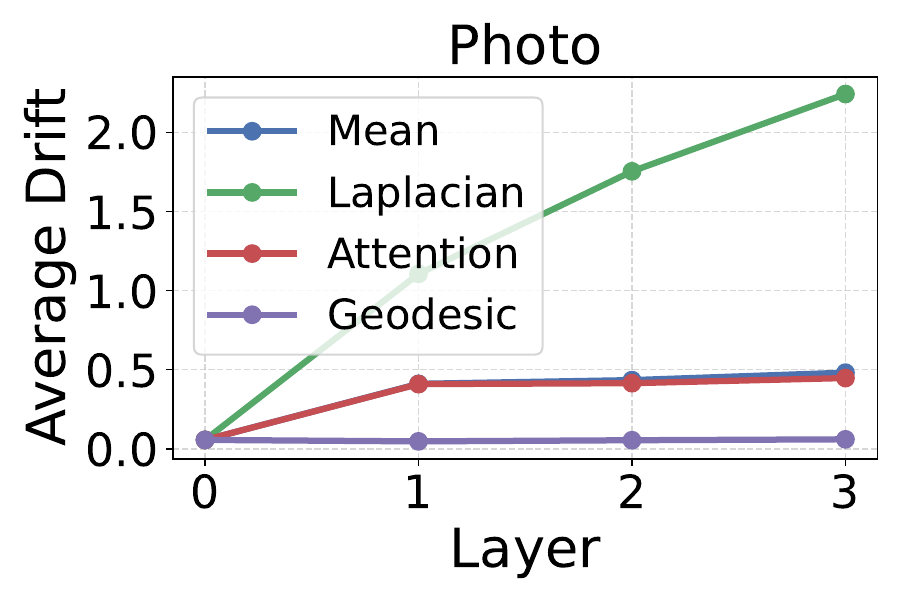}
  \includegraphics[width=0.485\linewidth]{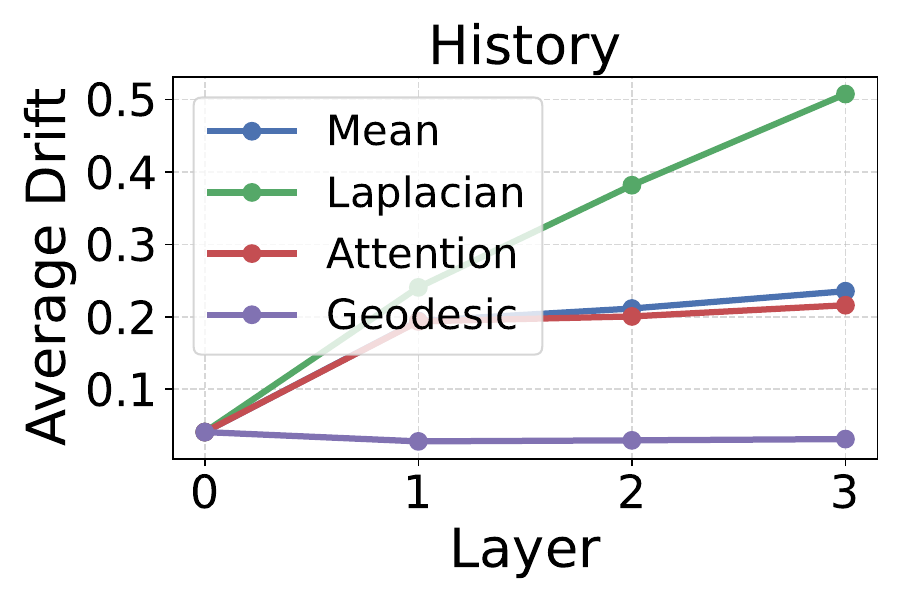}
  \caption{Quantifying semantic drift over aggregators.}
  \label{fig:geo_two}
\end{figure}

\subsection{Quantifying Semantic Drift}

While Definition~\ref{eq:semantic_drift} provides a conceptual view of semantic drift, 
directly computing geodesic deviation on an unknown manifold $\mathcal{M}$ 
is infeasible in practice. 
To approximate this deviation, we design a \textbf{local PCA-based metric} that 
quantifies how far each aggregated embedding has moved away from 
the locally estimated semantic manifold 
spanned by its PLM-based neighbors.

\noindent\textbf{Local Manifold Approximation.}
We assume that the feature space of a pretrained language model (PLM) 
forms a smooth semantic manifold in $\mathbb{R}^d$.
For each node $v_i$, we first identify its $k$ nearest neighbors 
in the PLM feature space 
$\mathbf{X}_{\text{PLM}} = [\mathbf{x}_1,\dots,\mathbf{x}_n]$ 
based on cosine similarity, forming a local neighborhood
$\mathcal{N}_i = \{\mathbf{x}_j \mid j \in \text{top-}k(i)\}$.
These neighbors characterize the local geometry of the semantic manifold
around $v_i$.

To estimate the local tangent subspace, we stack the neighbor embeddings
as a data matrix $\mathbf{X}_i \in \mathbb{R}^{k\times d}$ whose rows are the 
neighbor vectors, and compute their mean
$\bar{\mathbf{x}}_i = \frac{1}{k}\sum_{j\in \text{top-}k(i)}\mathbf{x}_j$.
After mean-centering, $\mathbf{Z}_i = \mathbf{X}_i - \mathbf{1}_k \bar{\mathbf{x}}_i^\top$,
we apply rank-$r$ Principal Component Analysis (PCA) to obtain the top-$r$ orthonormal
eigenvectors $\mathbf{V}_i^{(r)} \in \mathbb{R}^{d\times r}$ that span the local tangent subspace:
\begin{equation}
\mathcal{T}_i = \bar{\mathbf{x}}_i + \operatorname{span}\!\big(\mathbf{V}_i^{(r)}\big).
\label{eq:local_tangent}
\end{equation}
Intuitively, $\mathcal{T}_i$ serves as a linear approximation to the
semantic manifold near node $v_i$, capturing its locally dominant
directions of variation in the PLM embedding space.

\noindent\textbf{Reconstruction and Drift Measurement.}
Given the aggregated embedding $\mathbf{h}_i^{(l+1)}$ produced by the GNN, 
we assess how well it conforms to the local manifold estimated by PCA. 
Specifically, we first center it using the neighborhood mean $\bar{\mathbf{x}}_i$, 
apply PCA to obtain its projection onto the local tangent subspace $\mathcal{T}_i$, 
and reconstruct it back to the original space via the inverse PCA transform:
\begin{equation}
\tilde{\mathbf{h}}_i^{(l+1)} 
= \operatorname{PCA}^{-1}\!\big(\operatorname{PCA}(\mathbf{h}_i^{(l+1)} - \bar{\mathbf{x}}_i)\big) + \bar{\mathbf{x}}_i.
\end{equation}
This reconstruction is equivalent to orthogonally projecting 
$\mathbf{h}_i^{(l+1)}$ onto the affine subspace that locally approximates 
the PLM-induced semantic manifold.
The deviation between the original point and its reconstruction reflects 
how far $\mathbf{h}_i^{(l+1)}$ has moved off the local manifold, measured as:
\begin{equation}
E_i = 
\left\|
  \mathbf{h}_i^{(l+1)} - \tilde{\mathbf{h}}_i^{(l+1)}
\right\|_2^2,
\label{eq:local_pca_error}
\end{equation}
where $E_i$ denotes the local reconstruction error.  
To ensure scale invariance, we normalize this value by the squared norm of 
the centered embedding:
\begin{equation}
D_i = 
\frac{E_i}{\|\mathbf{h}_i^{(l+1)} - \bar{\mathbf{x}}_i\|_2^2 + \epsilon}.
\label{eq:normalized_drift}
\end{equation}
The mean drift score $\bar{D} = \frac{1}{|\mathcal{V}|}\sum_i D_i$
quantifies the overall extent to which a given aggregation mechanism 
distorts the underlying semantic manifold geometry.

\noindent\textbf{Neighborhood Reference.}
Importantly, the local tangent space $\mathcal{T}_i$ is always estimated 
using neighbors from the \emph{original PLM feature space}, 
rather than from the current GNN representations. 
This design anchors the drift measurement to the intrinsic semantic geometry 
learned by the pretrained encoder, providing a stable reference frame 
independent of the evolving GNN embeddings.
In other words, for each aggregated embedding $\mathbf{h}_i^{(l+1)}$, 
we ask whether it can still be well reconstructed from 
the manifold structure defined by its original semantic neighbors. 
A large reconstruction error thus directly reflects how much 
the aggregation has distorted the underlying semantic geometry.

\begin{figure*}[htbp]  % t=top, h=here, b=bottom, p=page of floats
    \centering
    \includegraphics[width=\linewidth]{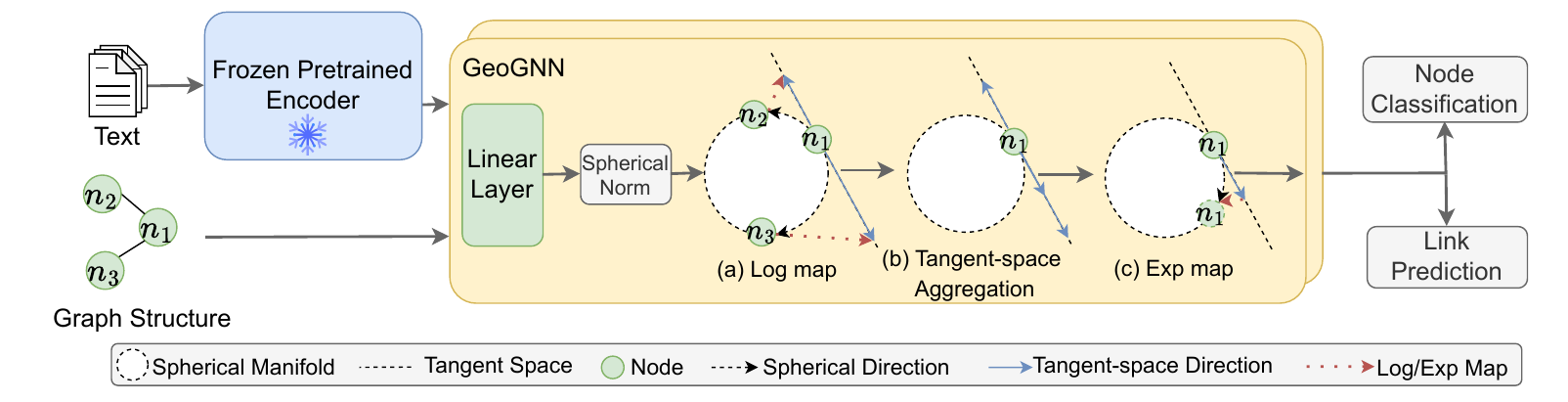}
\caption{
Overall framework of GeoGNN.
Node texts are encoded by a frozen pretrained language model (PLM) and projected onto a spherical manifold through linear projection and normalization. GeoGNN then performs geometry-preserving message passing by (a) mapping neighbor embeddings to the tangent space (\textit{log map}), (b) aggregating them via geodesic attention, and (c) projecting results back to the manifold (\textit{exp map}). This design preserves representation manifold fidelity.
}
    \label{fig:framework}
\end{figure*}

\noindent\textbf{Why This Quantifies Semantic Drift.}
Under the manifold hypothesis~\cite{ethayarajh2019contextual}, 
text embeddings from PLMs reside on a smooth, low-dimensional manifold, 
whose local geometry can be linearly approximated by its tangent subspace.  
If the GNN aggregation preserves this geometry, 
the updated representation $\mathbf{h}_i^{(l+1)}$ 
should be well reconstructed by its local tangent subspace $\mathcal{T}_i$.  
A large reconstruction error $E_i$ indicates that 
$\mathbf{h}_i^{(l+1)}$ has deviated from the local manifold, 
thus providing a direct, quantitative measure of \textbf{semantic drift}.
This notion parallels the tangent reconstruction error in manifold learning 
and locally linear embedding (LLE)~\cite{roweis2000nonlinear}, 
which also measures deviation from locally linear manifolds.

\noindent\textbf{Implementation.}
In practice, for each node $v_i$, 
we find its top-$k$ neighbors in the PLM feature space using cosine distance, 
fit a rank-$r$ PCA, 
project its GNN-updated embedding $\mathbf{h}_i^{(l+1)}$ onto the PCA subspace, 
and compute the normalized reconstruction error (Eqs.~\ref{eq:local_pca_error}–\ref{eq:normalized_drift}). 
This process can be efficiently implemented using 
\texttt{scikit-learn}'s \texttt{PCA} and parallelized across nodes.  
A higher $D_i$ indicates that the node’s updated representation 
lies farther from its semantic manifold, 
revealing a stronger geometric distortion introduced by the aggregation step.

\subsection{Empirical Evidence of Semantic Drift}
\label{sec:drift}

To visually and quantitatively validate the existence of semantic drift, 
we apply the proposed local PCA-based metric to embeddings produced by 
different aggregation mechanisms on two representative datasets: 
\textbf{Photo} and \textbf{History}. 
Figure~\ref{fig:agg_comparison} illustrates the local geometric structure of 
PLM-based node embeddings after one layer of aggregation using four common 
schemes: mean, Laplacian, attention, and our geodesic variant.

Aggregators such as mean and Attention clearly distort the original 
semantic geometry, flattening the curved manifold and pulling node embeddings 
toward the Euclidean mean of their neighbors. 
Laplacian-based aggregation exhibits a milder degree of distortion but still 
fails to align with the underlying manifold curvature. 
In contrast, the proposed \textbf{Geodesic Aggregation} preserves both 
the global curvature and local semantic continuity of the PLM-induced 
manifold, demonstrating high manifold fidelity and minimal geometric drift. 
These visualizations directly confirm that traditional linear message passing 
inevitably introduces semantic drift, while our geodesic design effectively 
maintains the intrinsic structure of the latent semantic space.

Beyond static visualization, we further quantify how semantic drift evolves 
with network depth. 
For each layer $l$, we compute the mean drift score $\bar{D}^{(l)}$ 
(Eq.~\ref{eq:normalized_drift}) for different aggregators. 
Figure~\ref{fig:geo_two} plots $\bar{D}^{(l)}$ across four layers on 
the Photo and History datasets. 
Linear aggregators exhibit rapidly increasing drift as layers deepen, 
reflecting severe cumulative manifold distortion. 
In contrast, Geodesic Aggregation maintains consistently low drift across 
all layers, confirming that manifold-aware message passing effectively 
mitigates semantic drift during propagation.
These empirical observations provide direct evidence that 
standard linear aggregation induces substantial semantic drift 
in text-attributed graphs. 
This motivates the design of our \textbf{Geodesic Aggregation}, 
a manifold-aware message passing mechanism that explicitly respects 
the geometry of the PLM-induced semantic manifold.

\section{Method}
\label{sec:method}
We present \textbf{GeoGNN} that performs message passing directly along geodesics on the semantic manifold induced by pretrained language models (PLMs). Each layer integrates three components: a learnable feature transformation, manifold-based aggregation via log–exp mappings, and multi-head geodesic attention.
An overview of the proposed framework is shown in Fig.~\ref{fig:framework}.
Given a text-attributed graph, node texts are first encoded by a frozen pretrained language model (PLM)
and then normalized onto a spherical manifold.
GeoGNN then performs geometry-preserving message passing via log–exp operations, as detailed below.

\subsection{Geodesic Aggregation Layer}

\noindent\textbf{1. Feature Transformation.}
Given node embeddings $\mathbf{h}_i^{(l)} \in \mathbb{R}^{d_l}$, 
we first apply a learnable linear projection to produce per-head representations:
\begin{equation}
\mathbf{z}_i^{(l)} = 
\operatorname{reshape}\!\big(
  \mathbf{W}^{(l)} \mathbf{h}_i^{(l)}
\big) \in \mathbb{R}^{H \times d_h},
\end{equation}
where $\mathbf{W}^{(l)} \in \mathbb{R}^{(d_{l+1}H) \times d_l}$ is trainable, 
$H$ denotes the number of attention heads, and $d_h=d_{l+1}$ is the per-head dimension.
This projection defines the learnable parameters of each GeoGNN layer, allowing 
different heads to capture distinct local geometries.

\noindent\textbf{2. Manifold Normalization.}
The latent feature space encoded by pretrained language models is highly non-linear and curved, 
making its intrinsic manifold difficult to explicitly control or estimate during training. 
To provide a stable and geometry-preserving proxy, 
we constrain all projected node embeddings to lie on a fixed and analytically tractable Riemannian manifold—the unit hypersphere $\mathbb{S}^{d_h-1}$. 
For head $h$, the normalized feature is computed as
\begin{equation}
\mathbf{x}_i^{(l,h)} = 
\frac{\mathbf{z}_i^{(l,h)}}{\|\mathbf{z}_i^{(l,h)}\|_2},
\quad 
\mathbf{x}_i^{(l,h)} \in \mathbb{S}^{d_h-1},
\label{eq:normalize}
\end{equation}
where the superscript $(h)$ denotes the $h$-th head.
We omit $(h)$ for brevity when not ambiguous.
This spherical normalization serves as an intrinsic coordinate system 
for subsequent geodesic operations, ensuring consistency and stability 
across layers while approximating the curvature of the PLM-induced semantic manifold.

\noindent\textbf{3. Log Map and Tangent-Space Aggregation.}
After constraining all features onto the unit hypersphere, 
we can explicitly operate on its well-defined Riemannian geometry.  
For each edge $(i,j)$, we map the neighbor representation 
$\mathbf{x}_j^{(l)}$ from the spherical manifold 
to the tangent space at the anchor node $\mathbf{x}_i^{(l)}$ 
using the logarithmic map:
\begin{equation}
\mathbf{v}_{ij} = 
\operatorname{Log}_{\mathbf{x}_i^{(l)}}(\mathbf{x}_j^{(l)})
=
\frac{\theta_{ij}}{\sin\theta_{ij}}
(\mathbf{x}_j^{(l)} - \cos\theta_{ij}\mathbf{x}_i^{(l)}),
\end{equation}
where 
$\theta_{ij} = \arccos((\mathbf{x}_i^{(l)})^\top \mathbf{x}_j^{(l)}) \in [0,\pi]$
is the geodesic distance between the two points on the sphere.
The logarithmic map yields a tangent vector 
$\mathbf{v}_{ij} \in T_{\mathbf{x}_i^{(l)}}\mathbb{S}^{d_h-1}$, 
the tangent space of the unit hypersphere at $\mathbf{x}_i^{(l)}$, 
which contains all vectors orthogonal to $\mathbf{x}_i^{(l)}$. 
Geometrically, $\mathbf{v}_{ij}$ represents the direction and magnitude 
of the shortest geodesic displacement from $\mathbf{x}_i^{(l)}$ to $\mathbf{x}_j^{(l)}$ 
on the manifold.

This formulation is computationally efficient because the unit sphere 
has a closed-form Riemannian metric and analytical log–exp mappings, 
allowing direct aggregation in tangent spaces without numerical approximation.  
Within each tangent space, we aggregate neighbors 
using a controlled geodesic attention mechanism:
\begin{equation}
a_{ij} = 
\frac{\exp(\cos\theta_{ij}/\tau)}
{\sum_{k\in\mathcal{N}(i)}\exp(\cos\theta_{ik}/\tau)},
\qquad
\mathbf{u}_i = \sum_{j\in\mathcal{N}(i)} a_{ij}\mathbf{v}_{ij},
\label{eq:tangent_agg}
\end{equation}
where $\tau$ is a fixed temperature hyperparameter controlling 
how sharply the model focuses on more semantically aligned neighbors.

Unlike conventional GNNs, where the attention weights 
or adjacency coefficients are learned implicitly through task-specific loss functions,
our design grounds neighborhood weighting directly in the 
\emph{semantic geometry} of the encoder’s feature space.
Specifically, we use the cosine similarity 
$\cos\theta_{ij} = (\mathbf{x}_i^{(l)})^\top \mathbf{x}_j^{(l)}$
as the measure of geodesic proximity, 
since it naturally corresponds to spherical distance on the manifold 
and has long been regarded as a faithful measure of semantic relatedness 
in NLP embeddings.
This choice eliminates the need for an additional learned distance metric,
preserving the intrinsic geometry and interpretability of the underlying 
semantic manifold.
The resulting vector $\mathbf{u}_i$ thus captures the aggregated 
semantic direction of neighborhood information 
in the tangent space at node $v_i$.

\noindent\textbf{4. Exponential Map Back to the Manifold.}
After aggregation in the tangent space, 
the resulting vector $\mathbf{u}_i$ is mapped back onto the manifold
via the \emph{exponential map}, 
which maps a tangent vector at $\mathbf{x}_i^{(l)}$
to another point on the manifold 
along the geodesic defined by that direction.
Formally, for a Riemannian manifold $\mathcal{M}$, 
$\operatorname{Exp}_{\mathbf{x}}(\mathbf{v})$ 
returns the endpoint of the geodesic starting from $\mathbf{x}$ 
with initial velocity $\mathbf{v}$ and length $\|\mathbf{v}\|$.

In our case, since the manifold is the unit hypersphere 
$\mathbb{S}^{d_h-1}$, 
the exponential map admits a closed-form expression:
\begin{equation}
\mathbf{x}_i^{(l+1)} 
= \operatorname{Exp}_{\mathbf{x}_i^{(l)}}(\alpha \mathbf{u}_i)
= 
\cos(\|\alpha\mathbf{u}_i\|)\mathbf{x}_i^{(l)}
+ 
\sin(\|\alpha\mathbf{u}_i\|)\frac{\mathbf{u}_i}{\|\mathbf{u}_i\|},
\end{equation}
where $\alpha$ is a fixed geodesic step-size hyperparameter 
that controls the magnitude of update along the manifold.
Intuitively, this operation follows the shortest curved path 
on the semantic manifold rather than a straight line in Euclidean space,
ensuring that the updated representation $\mathbf{x}_i^{(l+1)}$ 
remains on the sphere. 
This step completes one full geodesic message-passing cycle 
and ensures that semantic geometry is preserved throughout all layers.

\subsection{Multi-Head Attention and Layer Stacking}

Each head performs aggregation as in Eq.~\ref{eq:tangent_agg} with its own parameters. 
The resulting head outputs are concatenated as:
\begin{equation}
\mathbf{h}_i^{(l+1)} = 
\operatorname{Concat}\!\big(
\mathbf{x}_i^{(l+1,1)},\dots,\mathbf{x}_i^{(l+1,H)}
\big).
\end{equation}
Stacking multiple such layers yields the complete \textbf{GeoGNN}:
\begin{equation}
\mathbf{H}^{(L)} 
= 
\operatorname{GeoGNN}^{(L)}(\mathbf{X}, \mathcal{E}),
\end{equation}
where $\mathbf{X}$ are the PLM-derived node features and
$\mathcal{E}$ denotes the graph edges.
The final layer outputs $\mathbf{H}^{(L)}$ are used for tasks.

\subsection{Relation to Geodesic Smoothing}
In Section~\ref{sec:drift}, we introduced 
\textbf{Geodesic Smoothing} as a parameter-free variant of 
manifold-aware aggregation, used for analyzing and visualizing 
semantic drift across different aggregation mechanisms.
That version performs neighbor aggregation purely along geodesics 
via the same log–exp operations, but without any trainable parameters 
or attention weighting.
\textbf{GeoGNN}, introduced in this section, generalizes this idea
into a fully learnable framework.
It augments the parameter-free Geodesic Smoothing with:
(i) a linear projection, 
(ii) temperature-controlled geodesic attention, and 
(iii) a tunable geodesic step size as a hyperparameter.
Hence, Geodesic Smoothing serves as a non-parametric baseline 
for studying geometric behavior, while GeoGNN extends it into
an end-to-end trainable architecture for text-attributed graph learning.

\begin{table*}[htbp]
\centering
\caption{Node classification results on all datasets. Best results are in \textbf{bold}, second best are \underline{underlined}.}
\resizebox{0.9\textwidth}{!}{
\begin{tabular}{llcccccccccc}
\toprule
Dataset & Model & GCN & GAT & SAGE & GIN & SGC & RevGAT & APPNP & MLP & JKNet & GeoGNN \\
\midrule
\multirow{4}{*}{Photo} 
& Roberta\_base & 0.8230 & 0.8248 & 0.8305 & 0.7523 & 0.7621 & \underline{0.8321} & 0.8188 & 0.6481 & 0.7770 & \textbf{0.8438} \\
& MiniLM-L6-v2 & 0.8390 & 0.8402 & \underline{0.8450} & 0.7445 & 0.8036 & 0.8429 & 0.8309 & 0.6932 & 0.8146 & \textbf{0.8499} \\
% & Distilbert\_base & 0.8210 & \underline{0.8336} & 0.8333 & 0.7168 & 0.7332 & 0.8291 & 0.8073 & 0.6420 & 0.7955 & \textbf{0.8451} \\
& Bert\_base & 0.8226 & 0.8308 & 0.8311 & 0.7037 & 0.7418 & \underline{0.8256} & 0.8025 & 0.6344 & 0.7932 & \textbf{0.8445} \\
% & Roberta\_large & 0.8291 & 0.8238 & \underline{0.8374} & 0.6807 & 0.5805 & 0.8384 & 0.8241 & 0.6431 & 0.8097 & \textbf{0.8492} \\
% & Bert\_large & 0.8178 & \textbf{0.8354} & \underline{0.8294} & 0.6935 & 0.7335 & 0.8254 & 0.8112 & 0.6413 & 0.7876 & 0.8405 \\
& Sentence-t5-large & 0.8395 & 0.8422 & \underline{0.8462} & 0.7458 & 0.7909 & 0.8452 & 0.8323 & 0.7040 & 0.8137 & \textbf{0.8549} \\
\midrule
\multirow{4}{*}{Children} 
& Roberta\_base & 0.5692 & 0.5433 & 0.5725 & 0.5704 & 0.4534 & \underline{0.5740} & 0.5500 & 0.5452 & 0.4614 & \textbf{0.5897} \\
& MiniLM-L6-v2 & 0.5777 & 0.5731 & \underline{0.5972} & 0.5578 & 0.4700 & 0.5974 & 0.5780 & 0.5316 & 0.5391 & \textbf{0.6085} \\
% & Distilbert\_base & 0.5674 & 0.5628 & \underline{0.5988} & 0.5797 & 0.4284 & 0.5898 & 0.5466 & 0.5412 & 0.4970 & \textbf{0.6114} \\
& Bert\_base & 0.5696 & 0.5636 & \underline{0.5949} & 0.5775 & 0.4349 & 0.5917 & 0.5572 & 0.5479 & 0.5089 & \textbf{0.6146} \\
% & Roberta\_large & 0.5650 & 0.5353 & \underline{0.5587} & 0.5660 & 0.3600 & 0.5795 & 0.5491 & 0.4983 & 0.5004 & \textbf{0.5995} \\
% & Bert\_large & 0.5620 & 0.5571 & \underline{0.5918} & 0.5816 & 0.4300 & 0.5876 & 0.5464 & 0.5463 & 0.4795 & \textbf{0.5813} \\
& Sentence-t5-large & 0.5688 & 0.5715 & \underline{0.6068} & 0.5859 & 0.4556 & 0.6057 & 0.5611 & 0.5644 & 0.4916 & \textbf{0.6184} \\
\midrule
\multirow{4}{*}{Arxiv} 
& Roberta\_base\_512\_cls & 0.7256 & 0.7042 & 0.7349 & 0.7140 & 0.6701 & \underline{0.7335} & 0.7293 & 0.6840 & 0.6910 & \textbf{0.7454} \\
& MiniLM-L6-v2 & 0.7361 & 0.7335 & \underline{0.7493} & 0.7233 & 0.7039 & \underline{0.7521} & 0.7455 & 0.7226 & 0.7067 & \textbf{0.7556} \\
% & Distilbert\_base & 0.7325 & 0.7312 & \underline{0.7420} & 0.7164 & 0.6365 & 0.7400 & 0.7312 & 0.6831 & 0.7002 & \textbf{0.7478} \\
& Bert\_base & 0.7286 & 0.7285 & \underline{0.7439} & 0.7166 & 0.6462 & 0.7369 & 0.7316 & 0.6864 & 0.7010 & \textbf{0.7499} \\
% & Roberta\_large & 0.7346 & 0.7194 & \underline{0.7384} & 0.7117 & 0.4731 & 0.7406 & 0.7370 & 0.6959 & 0.7022 & \textbf{0.7568} \\
% & Bert\_large & 0.7283 & 0.7283 & \underline{0.7480} & 0.7192 & 0.6442 & 0.7449 & 0.7361 & 0.6990 & 0.7005 & \textbf{0.7522} \\
& sentence-t5-large & 0.7331 & 0.7277 & \underline{0.7439} & 0.7184 & 0.6819 & 0.7443 & 0.7375 & 0.7046 & 0.7050 & \textbf{0.7524} \\
\midrule
\multirow{4}{*}{History} 
& Roberta\_base & 0.8474 & 0.8339 & 0.8375 & 0.8405 & 0.8202 & 0.8479 & \underline{0.8514} & 0.8367 & 0.8271 & \textbf{0.8579} \\
& MiniLM-L6-v2 & 0.8467 & 0.8451 & 0.8527 & 0.8185 & 0.8285 & \underline{0.8528} & 0.8536 & 0.8328 & 0.8336 & \textbf{0.8617} \\
% & Distilbert\_base & 0.8476 & 0.8438 & 0.8529 & 0.8402 & 0.8135 & \textbf{0.8557} & \underline{0.8553} & 0.8372 & 0.8371 & 0.8630 \\
& Bert\_base & 0.8472 & 0.8437 & 0.8536 & 0.8371 & 0.8168 & \underline{0.8558} & 0.8552 & 0.8392 & 0.8375 & \textbf{0.8624} \\
% & Roberta\_large & 0.8452 & 0.8351 & 0.8418 & 0.8272 & 0.7492 & \underline{0.8515} & 0.8534 & 0.8216 & 0.8354 & \textbf{0.8591} \\
% & Bert\_large & 0.8478 & 0.8434 & 0.8486 & 0.8370 & 0.8123 & \underline{0.8542} & 0.8506 & 0.8414 & 0.8382 & \textbf{0.8581} \\
& Sentence-t5-large & 0.8472 & 0.8452 & \underline{0.8587} & 0.8328 & 0.8247 & 0.8585 & 0.8531 & 0.8439 & 0.8353 & \textbf{0.8648} \\
\bottomrule
\end{tabular}
}
\label{tab:all_results}
\end{table*}

\section{Experiments}
\label{sec:exp}

\subsection{Experimental Setup.}
We conduct experiments on four text-attributed graph datasets from the CS-TAG benchmark~\cite{yan2023cstag}: 
Photo, Children, Arxiv, and History, 
covering both e-commerce and academic domains. 
Node texts are encoded by frozen pretrained language models (PLMs) 
including RoBERTa~\cite{liu2019roberta}, BERT~\cite{devlin2019bert}, MiniLM~\cite{wang2020minilm}, and Sentence-T5~\cite{ni2021sentence}, 
whose embeddings serve as fixed node features for all graph models. 
We compare GeoGNN with nine representative GNN baselines 
(GCN~\cite{kipf2017semi}, GAT~\cite{velickovic2018graph}, GraphSAGE~\cite{hamilton2017inductive}, 
GIN~\cite{xu2018powerful}, SGC~\cite{wu2019simplifying}, RevGAT~\cite{li2021training}, 
APPNP~\cite{gasteiger2018predict}, JKNet~\cite{xu2018representation} and MLP) 
under identical settings for both node classification and link prediction tasks. 
All models are trained using the Adam optimizer on NVIDIA H200 GPUs; 
full dataset statistics, hyperparameter configurations, and implementation details 
are provided in Appendix~\ref{appendix:setup}.

\subsection{Node Classification and Link Prediction Experiment Results}

\noindent\textbf{Overall Results.}
Tables~\ref{tab:all_results} and~\ref{tab:link_prediction_full} summarize node classification and link prediction results on four benchmark datasets. 
GeoGNN consistently achieves the best scores across all settings, validating the effectiveness of geodesic aggregation in mitigating semantic drift. 
On average, GeoGNN improves node classification accuracy by 2–3\% and Hit@10 by 4–6\% over the strongest baselines, showing both higher performance and greater stability across encoders.

\begin{figure*}[htbp]
  \centering
  \includegraphics[width=0.245\linewidth]{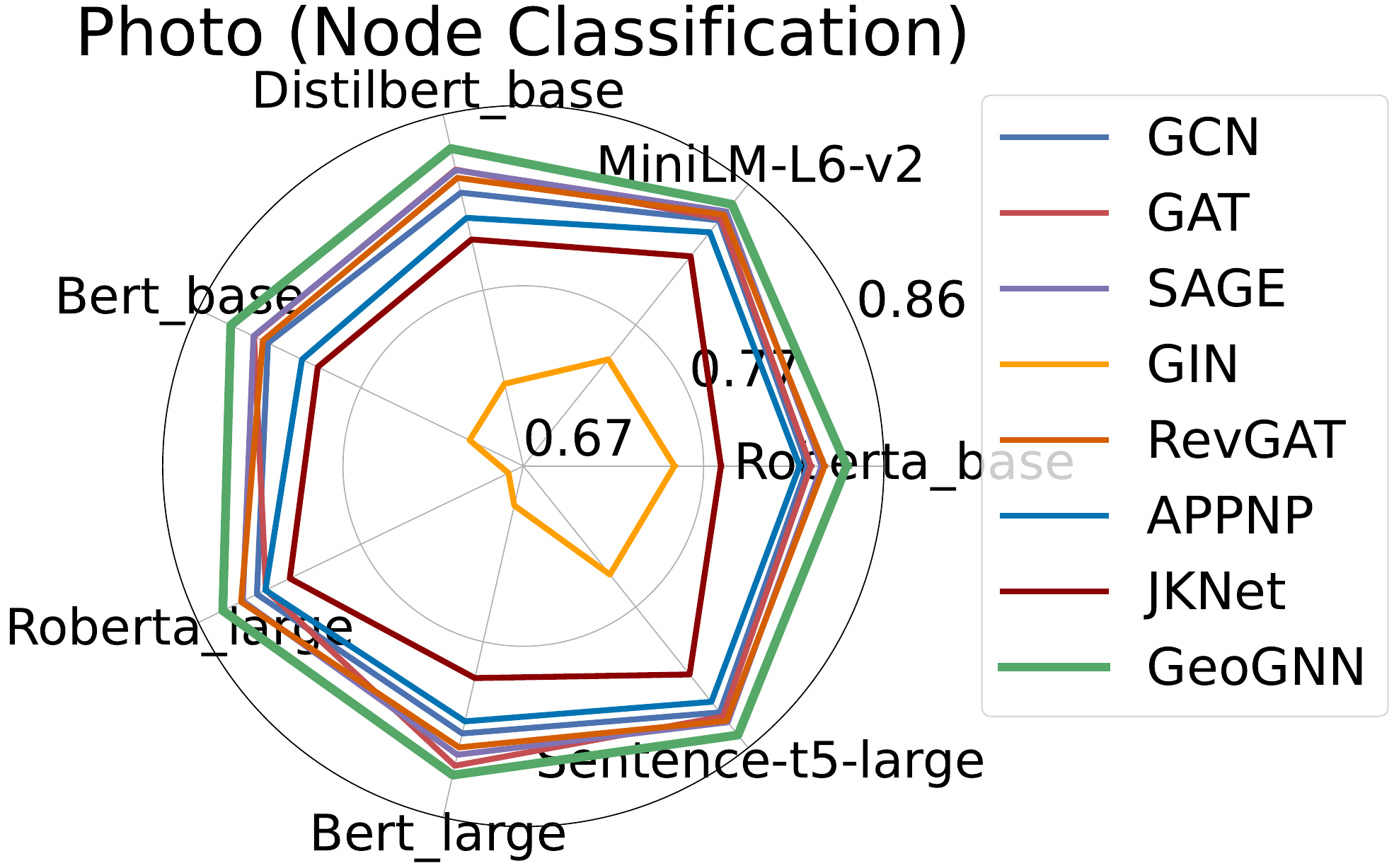}
  \includegraphics[width=0.245\linewidth]{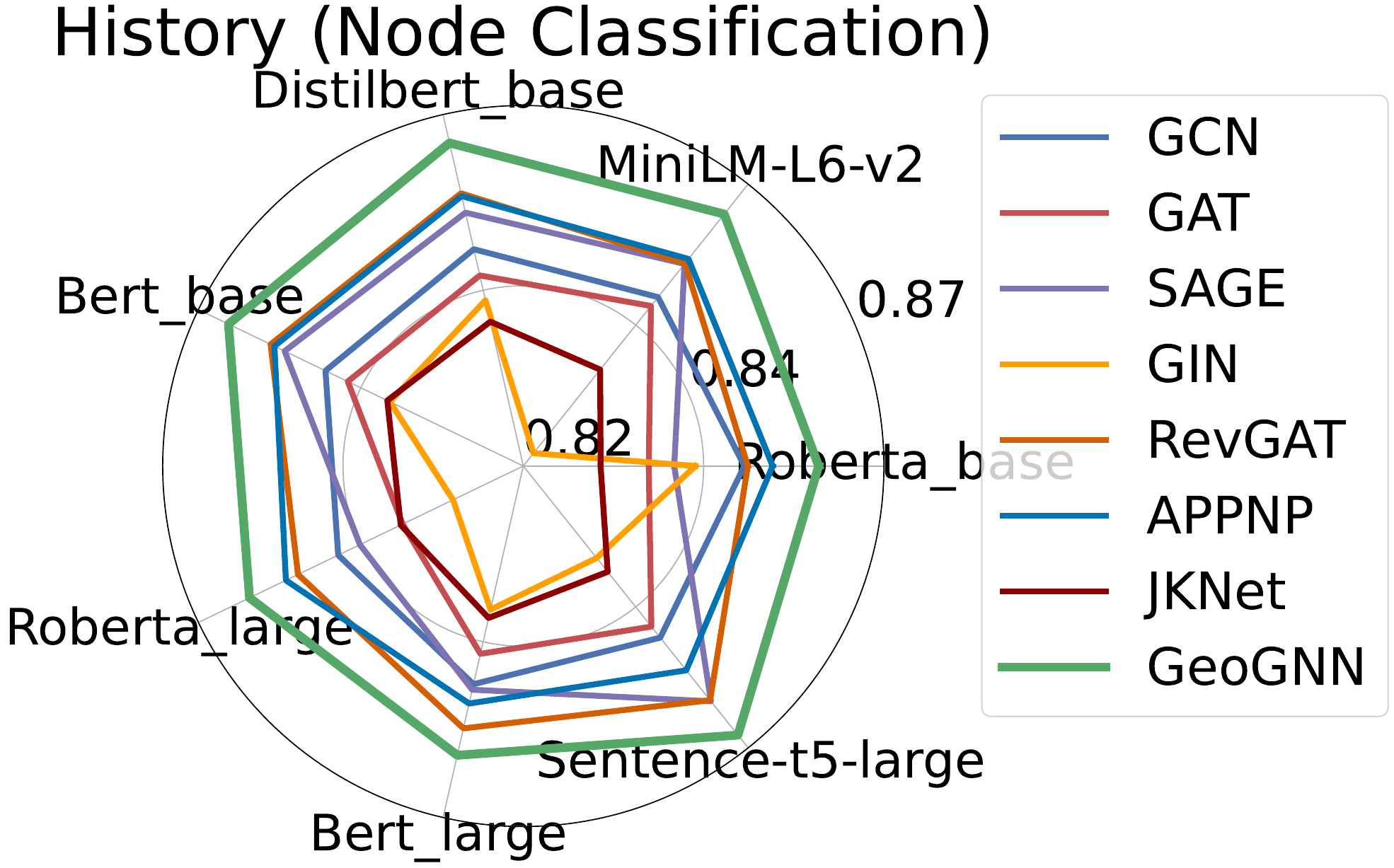}
  \includegraphics[width=0.245\linewidth]{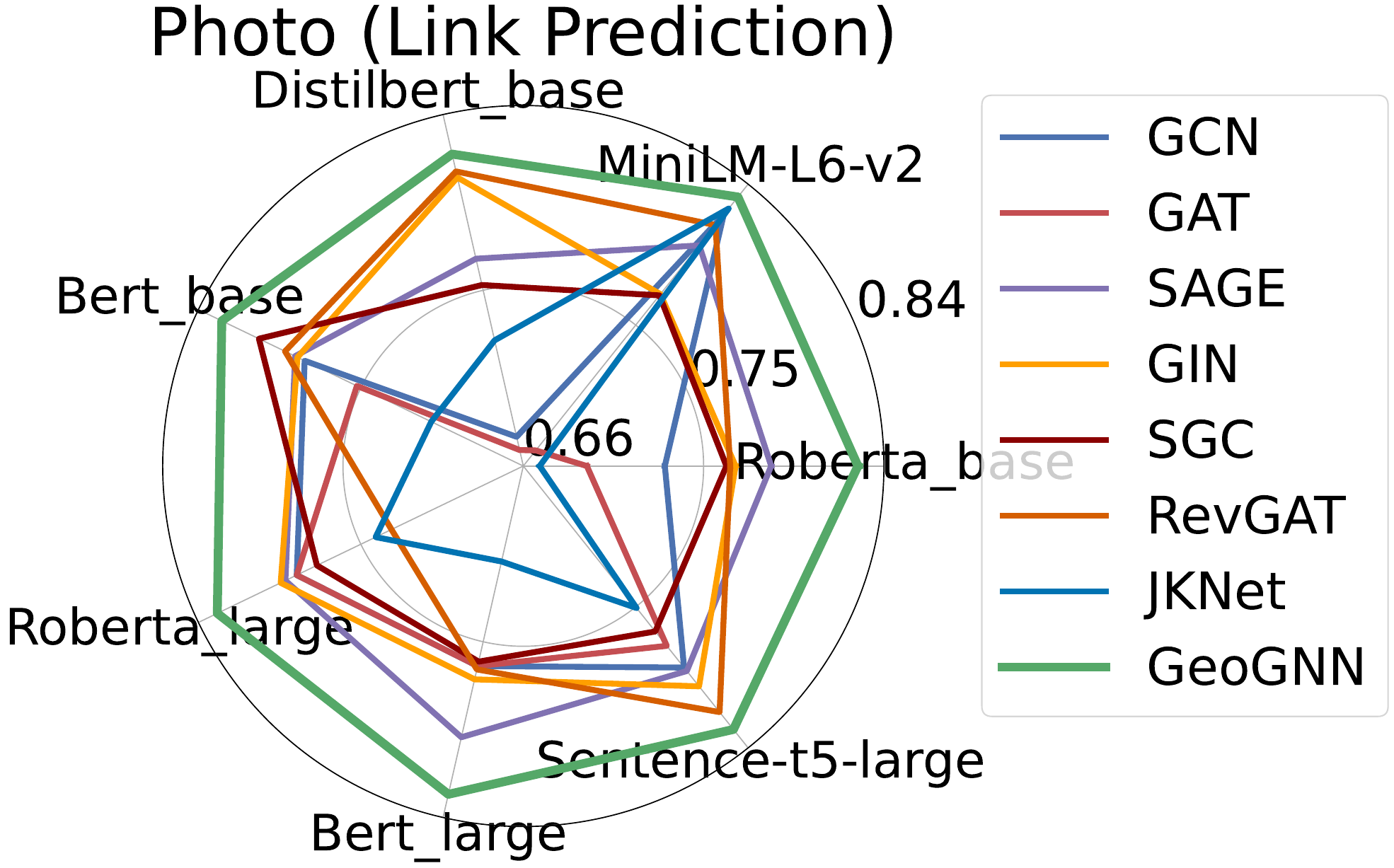}
  \includegraphics[width=0.245\linewidth]{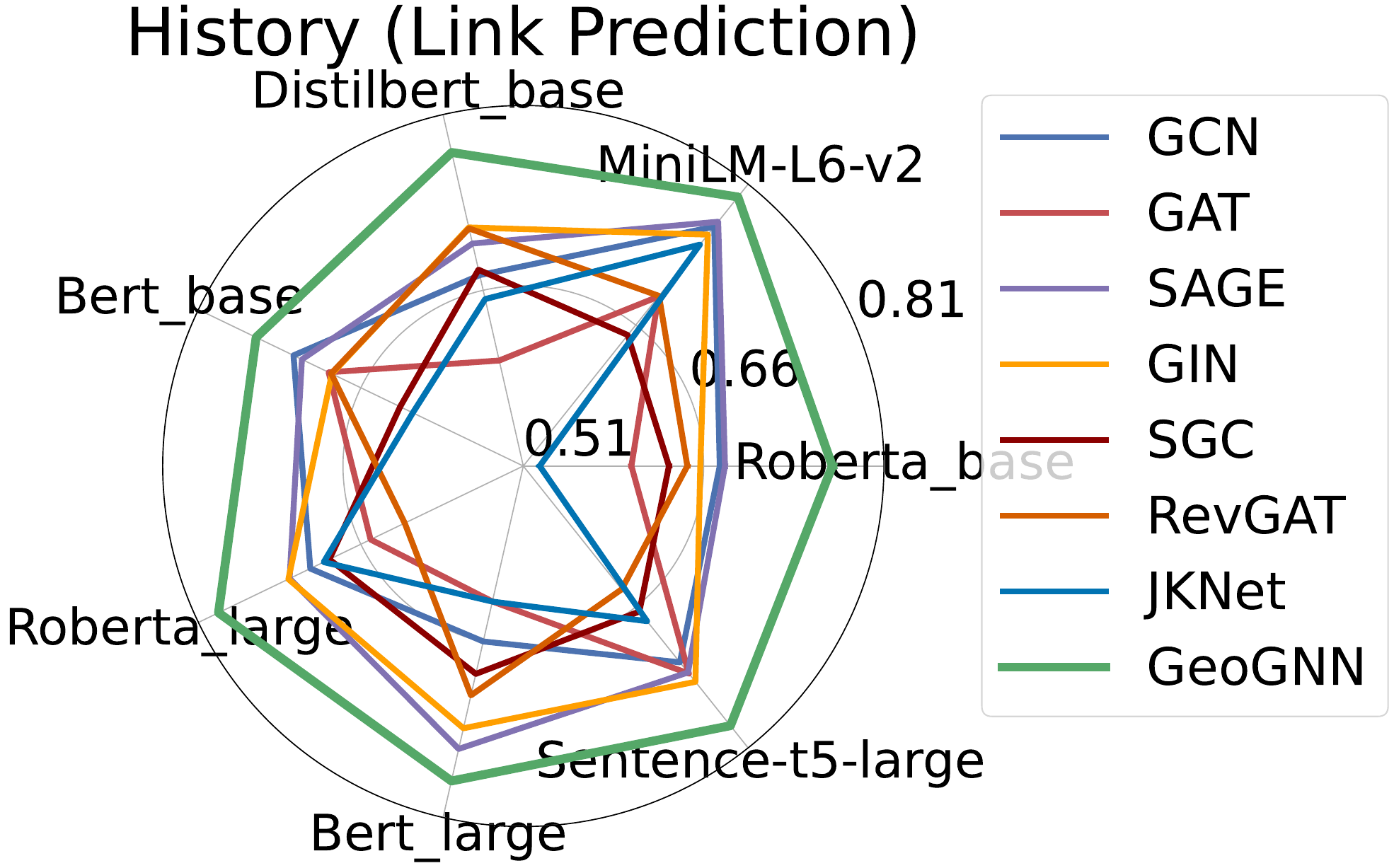}
  \caption{Comparison between GNNs over different encoders}
  \label{fig:compare_among_encoders}
\end{figure*}

\medskip
\noindent\textbf{Node Classification.}
As shown in Table~\ref{tab:all_results} and the radar plots for Photo and History (Figure~\ref{fig:compare_among_encoders}), GeoGNN consistently outperforms conventional GNNs under all pretrained text encoders.
The improvement is most pronounced on \textbf{Children} (+3.2\% on average) and \textbf{Photo} (+1.5\%), where node texts are noisy and semantically diverse—conditions under which linear aggregation easily causes semantic drift.
In contrast, GeoGNN preserves the manifold curvature of PLM embeddings, resulting in smoother yet discriminative node representations.
Notably, on \textbf{History}, GeoGNN reaches 0.86 accuracy with the lowest variance among all baselines, confirming its robustness to domain shift.
Models coupled with semantically rich encoders such as Sentence-T5 and RoBERTa further amplify GeoGNN’s advantage, indicating that manifold-aware message passing fully leverages the geometric structure of stronger PLMs.

\begin{table*}[htbp]
\centering
\caption{Link Prediction results (Hit@10) on all datasets. Best results are in \textbf{bold}, second best are \underline{underlined}.}
\resizebox{0.9\textwidth}{!}{
\begin{tabular}{llccccccccccc}
\toprule
Dataset & Model & GCN & GAT & SAGE & GIN & SGC & RevGAT & APPNP & MLP & JKNet & GeoGNN \\
\midrule
\multirow{4}{*}{Photo}
& Roberta\_base & 0.7301 & 0.6911 & \underline{0.7833} & 0.7655 & 0.7609 & 0.7628 & 0.3012 & 0.2871 & 0.6675 & \textbf{0.8267} \\
& MiniLM-L6-v2 & 0.8201 & 0.6693 & 0.8005 & 0.7693 & 0.7685 & 0.8135 & 0.7892 & 0.3684 & \underline{0.8241} & \textbf{0.8315} \\
& Bert\_base & 0.7807 & 0.7517 & 0.7860 & 0.7847 & \underline{0.8061} & 0.7916 & 0.4078 & 0.2812 & 0.7104 & \textbf{0.8267} \\
& Sentence-t5-large & 0.7883 & 0.7744 & 0.7905 & 0.8003 & 0.7652 & \underline{0.8168} & 0.1267 & 0.3673 & 0.7502 & \textbf{0.8279} \\
\midrule
\multirow{4}{*}{Children}
& Roberta\_base & 0.6824 & 0.6782 & 0.7229 & \underline{0.7503} & 0.7361 & 0.6694 & 0.1390 & 0.4627 & 0.3036 & \textbf{0.7748} \\
& MiniLM-L6-v2 & 0.7624 & 0.6619 & \underline{0.7669} & 0.7404 & 0.7158 & 0.6702 & 0.6524 & 0.4356 & 0.6640 & \textbf{0.7820} \\
& Bert\_base & 0.7168 & 0.6714 & 0.7239 & 0.6947 & \underline{0.7324} & 0.6939 & 0.1523 & 0.5457 & 0.5913 & \textbf{0.7479} \\
& Sentence-t5-large & 0.6621 & 0.6628 & \underline{0.7382} & 0.7298 & 0.7177 & 0.7378 & 0.1177 & 0.4892 & 0.5559 & \textbf{0.7635} \\
\midrule
\multirow{4}{*}{Arxiv}
& Roberta\_base & 0.6253 & 0.6166 & \underline{0.6261} & 0.5863 & 0.6223 & 0.5542 & 0.2515 & 0.4701 & 0.5617 & \textbf{0.6444} \\
& MiniLM-L6-v2 & 0.6268 & 0.6603 & 0.6100 & 0.6574 & 0.7136 & 0.3973 & 0.6569 & 0.5150 & \underline{0.7321} & \textbf{0.7387} \\
& Bert\_base & \underline{0.6766} & 0.6414 & 0.6712 & 0.6359 & 0.6717 & 0.5054 & 0.2780 & 0.4704 & 0.6032 & \textbf{0.6833} \\
& Sentence-t5-large & 0.6443 & 0.6387 & 0.6438 & 0.6128 & 0.6421 & 0.5215 & 0.2233 & 0.4938 & \underline{0.6622} & \textbf{0.6625} \\
\midrule
\multirow{4}{*}{History}
& Roberta\_base & 0.6708 & \underline{0.6978} & 0.6751 & 0.6547 & 0.6288 & 0.6440 & 0.4622 & 0.4512 & 0.5220 & \textbf{0.7638} \\
& MiniLM-L6-v2 & \underline{0.7615} & 0.6876 & \underline{0.7667} & 0.7531 & 0.6469 & 0.6883 & 0.6771 & 0.4351 & 0.7425 & \textbf{0.7929} \\
& Bert\_base & \underline{0.7192} & 0.6870 & 0.7115 & 0.6840 & 0.6214 & 0.6847 & 0.4685 & 0.5111 & 0.6098 & \textbf{0.7535} \\
& Sentence-t5-large & 0.7160 & 0.7279 & 0.7267 & \underline{0.7366} & 0.6623 & 0.6388 & 0.3391 & 0.5081 & 0.6725 & \textbf{0.7830} \\
\bottomrule
\end{tabular}
}
\label{tab:link_prediction_full}
\end{table*}

\medskip
\noindent\textbf{Link Prediction.}
Table~\ref{tab:link_prediction_full} and the radar plots for Photo and History (Figure~\ref{fig:compare_among_encoders}) reveal even clearer trends.
GeoGNN consistently achieves the highest Hit@10 on all datasets, surpassing the best baseline by 5.8\% on average. 
In particular, on \textbf{Photo} and \textbf{Children}, GeoGNN attains 0.79–0.83 Hit@10, substantially ahead of the next best models. 
This demonstrates that geodesic message passing preserves local semantic topology crucial for accurate edge-level similarity estimation. 
Linear aggregation methods (e.g., GCN, SGC) collapse semantic neighborhoods into Euclidean averages, leading to under-discriminative link scores, while GeoGNN follows the true manifold geodesics to maintain fine-grained relational geometry.

\medskip
\noindent\textbf{Summary.}
Across both node- and edge-level tasks, GeoGNN consistently improves performance and stability under different encoders and datasets. 
These results strongly support our central hypothesis: \emph{semantic drift arising from linear aggregation is the key bottleneck in text-attributed graphs, and aligning message passing with manifold geometry provides a principled and empirically validated solution.}

\subsection{Hyperparameter Sensitivity}

\begin{figure*}[htbp]
  \centering
  \includegraphics[width=0.245\linewidth]{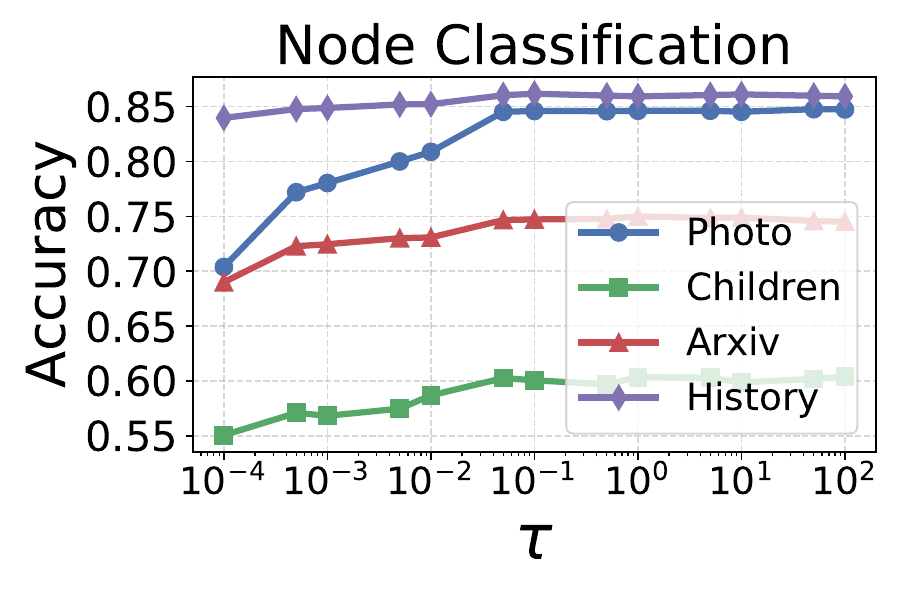}
  \includegraphics[width=0.245\linewidth]{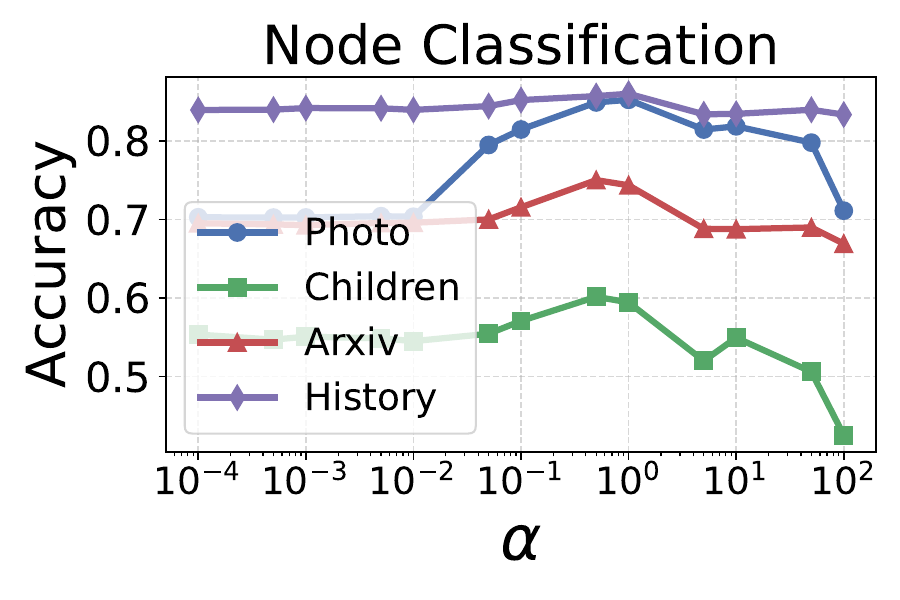}
  \includegraphics[width=0.245\linewidth]{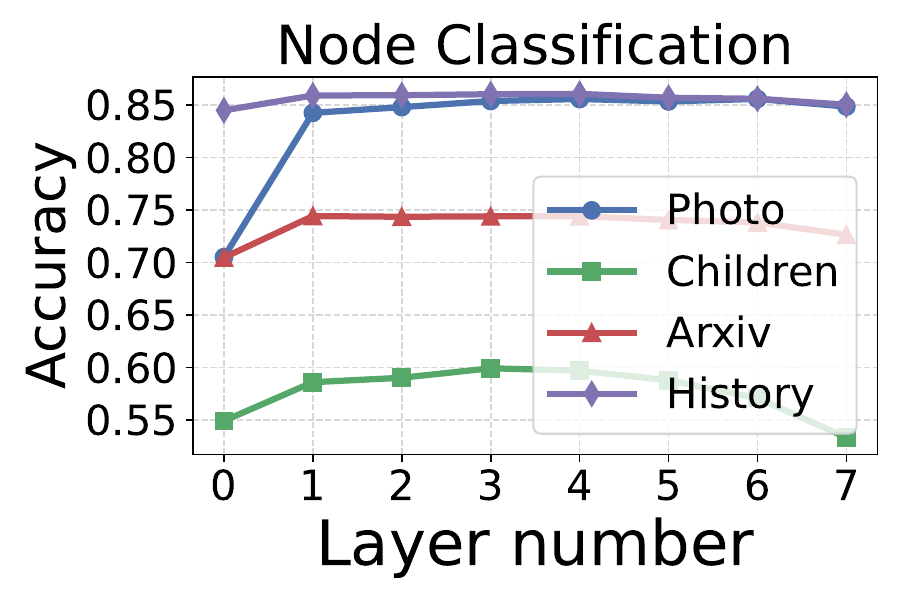}
  \includegraphics[width=0.245\linewidth]{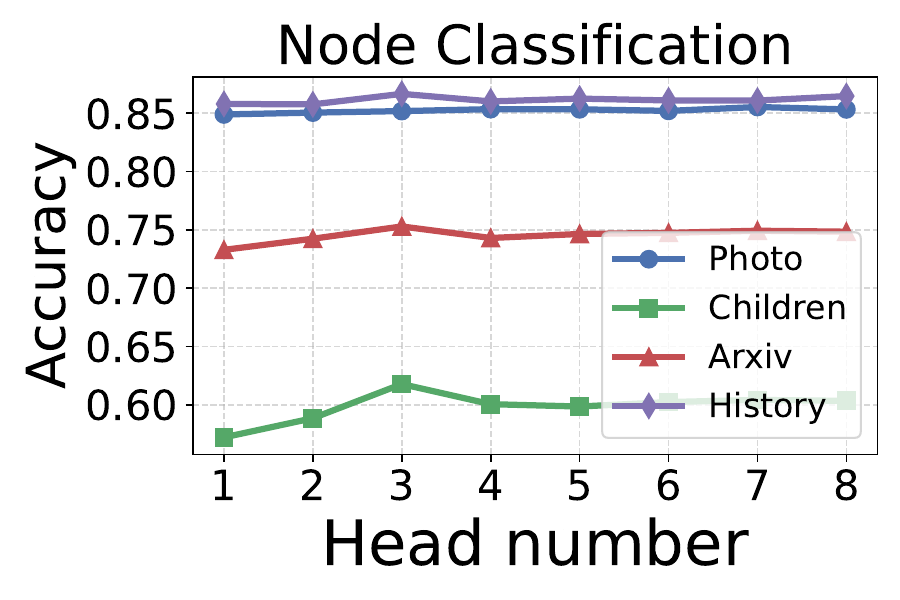}
  \includegraphics[width=0.245\linewidth]{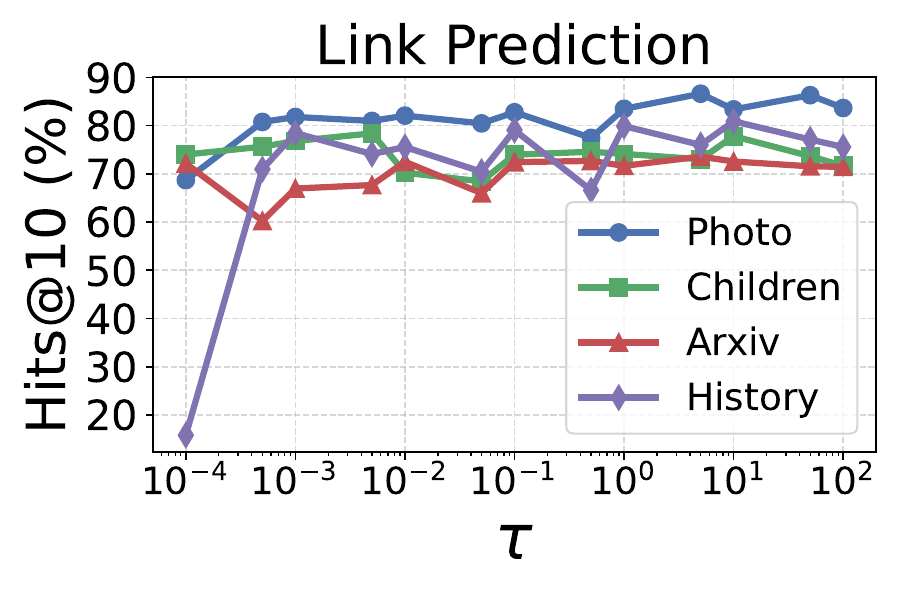}
  \includegraphics[width=0.245\linewidth]{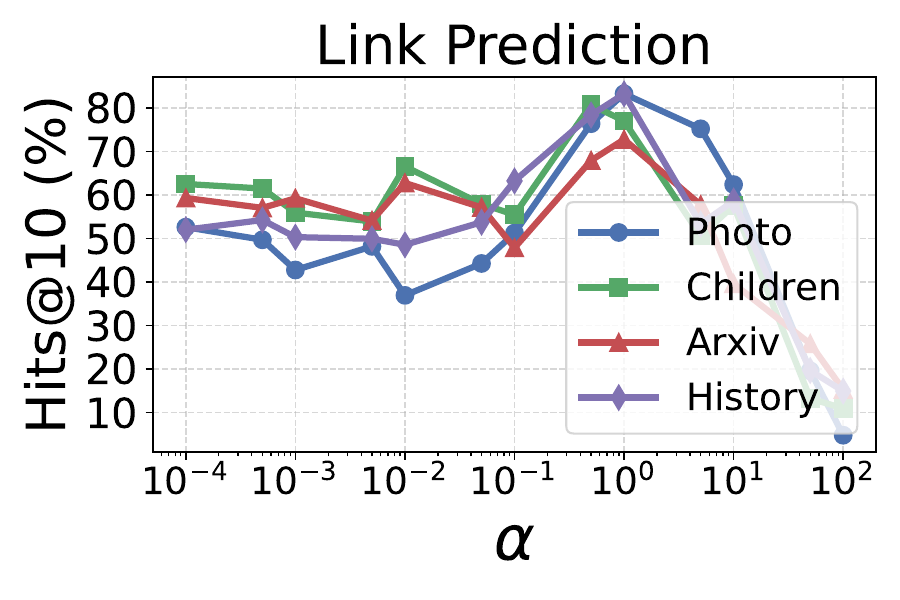}
  \includegraphics[width=0.245\linewidth]{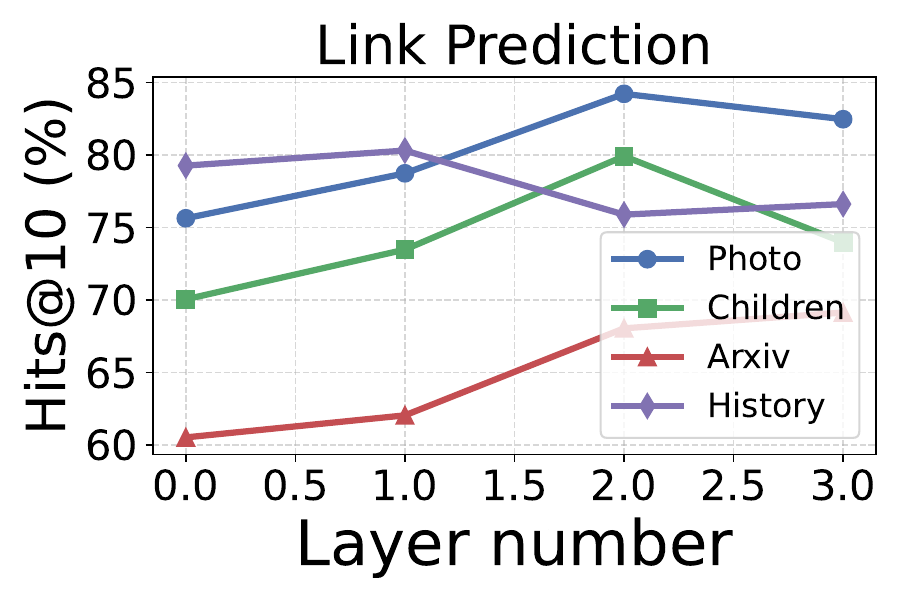}
  \includegraphics[width=0.245\linewidth]{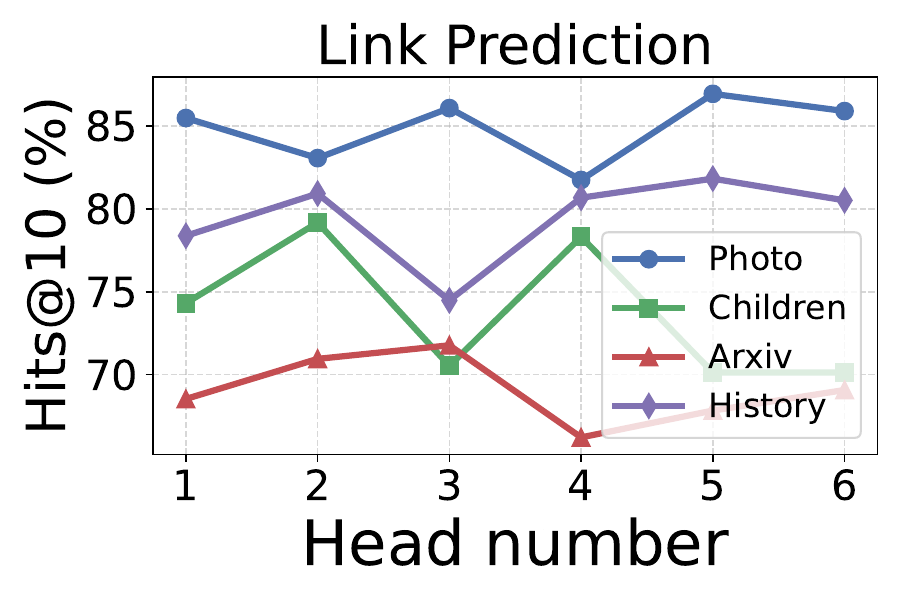}  
  \caption{Hyperparameter sensitivity}
  \label{fig:hyperparam}
\end{figure*}

We further analyze the effect of four key hyperparameters: the \textbf{temperature} $\tau$, \textbf{geodesic step size} $\alpha$, \textbf{number of GNN layers}, and \textbf{number of attention heads} --- on both node classification and link prediction tasks, as illustrated in Figure~\ref{fig:hyperparam}.

\textbf{Temperature $\tau$.} 
As $\tau$ controls the sharpness of geodesic distance scaling, performance first improves and then stabilizes as $\tau$ increases. 
Extremely small values lead to under-scaled distances and unstable gradients, while overly large values cause over-smoothing of node embeddings. 
For both tasks, $\tau$ in the range of \textbf{1--10} yields the best results, indicating that moderate geometric temperature preserves both local and global relational information.

\textbf{Geodesic step size $\alpha$.} 
The parameter $\alpha$ determines the magnitude of manifold updates. 
When $\alpha$ is too small, the geodesic propagation becomes ineffective, leading to weak structure modeling; 
when $\alpha$ is too large, overshooting occurs, resulting in curvature distortion and unstable optimization. 
In both node and link settings, performance peaks around \textbf{$\alpha \approx 0.5$--$1$}, where the model balances expressiveness and stability.

\textbf{Number of layers.} 
The performance curves exhibit the typical GNN trend: accuracy improves with depth up to \textbf{2--3 layers} and then decreases due to over-smoothing. 
This observation suggests that moderate propagation depth is sufficient for effective geodesic message passing without excessive feature mixing.

\textbf{Number of attention heads.} 
Increasing the number of heads initially enhances representation diversity and stabilizes training by enabling multi-subspace aggregation. 
However, beyond \textbf{6--8 heads}, the improvement saturates or slightly declines, likely due to redundant attention patterns and higher variance. 
This consistent trend across datasets shows that the multi-head design in \textit{GeoGNN} is robust yet does not require excessive parameterization.

Overall, \textit{GeoGNN} demonstrates strong robustness to hyperparameters. 
The model maintains stable performance across a wide range of $\tau$ and $\alpha$, as well as architectural parameters, underscoring its reliable geometric design and generalizability across tasks.

\subsection{Ablation Study}

To understand the contribution of each geometric component in GeoGNN, 
we conduct ablation studies using three representative encoders of different sizes
(\textit{MiniLM-L6-v2}, \textit{BERT-base}, and \textit{Sentence-T5-large}). 
We compare the full GeoGNN with three simplified variants:
(1) \textbf{w/o Geodesic} — replacing log–exp mappings with linear aggregation, 
(2) \textbf{w/o Cos} — removing cosine-based geodesic attention and using uniform neighbor weighting, and
(3) \textbf{w/o Normalization} — disabling spherical normalization of feature vectors.

\begin{figure}[htbp]
  \centering
  \includegraphics[width=0.485\linewidth]{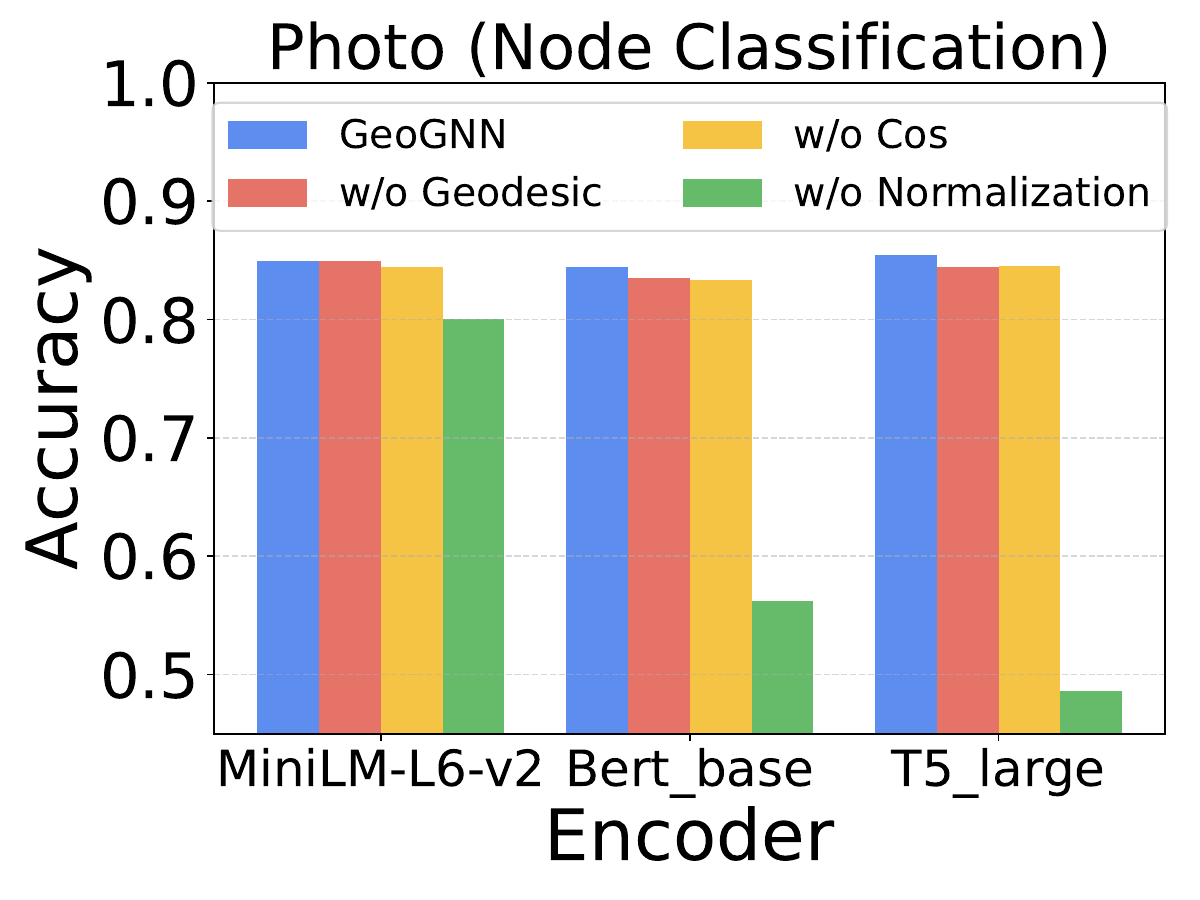}
 \includegraphics[width=0.485\linewidth]{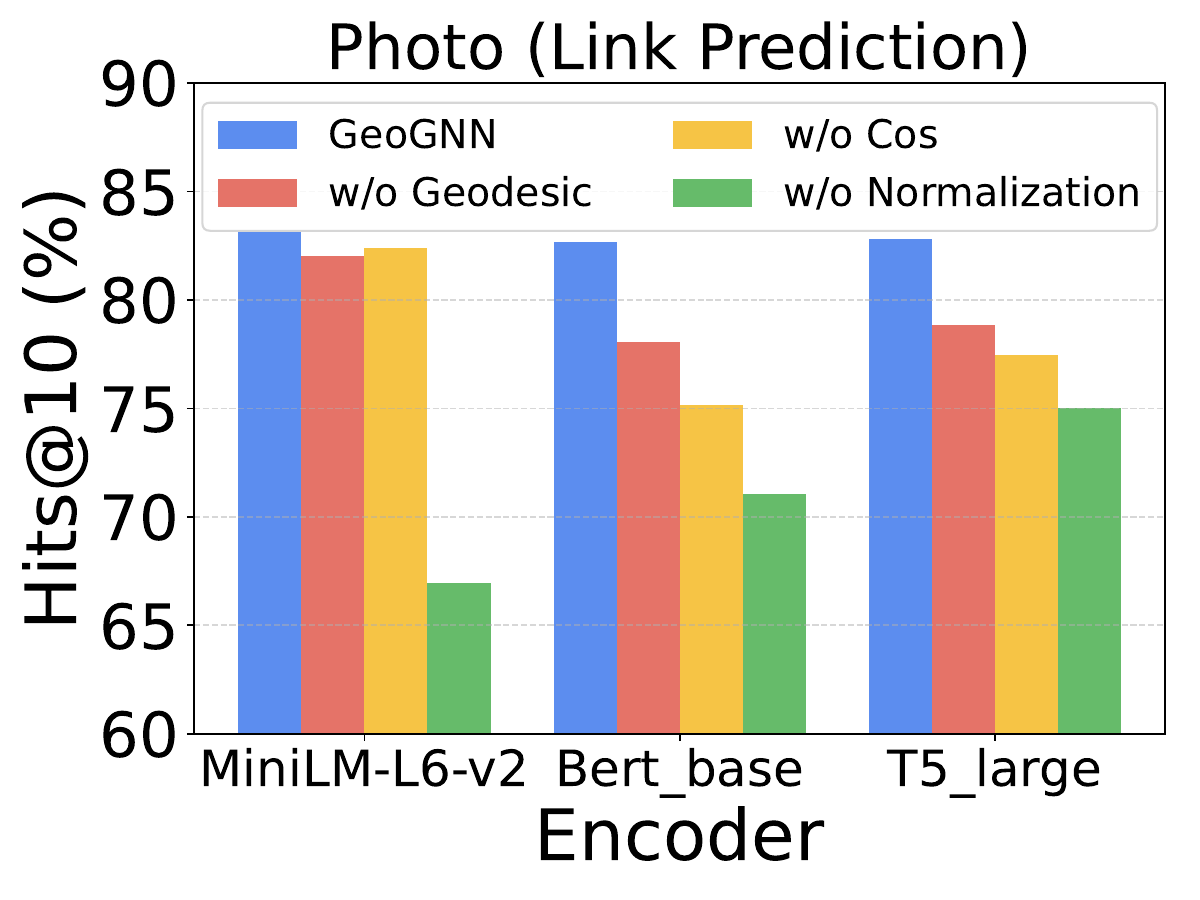}
  \caption{Ablation Study}
  \label{fig:ablation}
\end{figure}

\medskip
\noindent\textbf{Results and Analysis.}
Figure~\ref{fig:ablation} summarizes the results for node classification (left) 
and link prediction (right) on the Photo dataset.
The complete GeoGNN consistently achieves the best performance across all encoders.
Removing geodesic aggregation (\textit{w/o Geodesic}) causes the largest degradation 
($-3.0\%$ on average for node classification and $-4.5\%$ for link prediction), 
demonstrating that following the manifold geodesics rather than linear interpolation 
is critical to preserving semantic geometry. 
Eliminating spherical normalization (\textit{w/o Normalization}) slightly reduces accuracy 
but leads to unstable convergence, as the manifold constraint is necessary 
to maintain consistent curvature across layers. 
Finally, removing cosine-based weighting (\textit{w/o Cos}) harms models with 
semantically rich encoders (e.g., Sentence-T5), highlighting that 
geodesic attention effectively aligns neighborhood weighting 
with intrinsic semantic similarity.
We observe similar trends on all the other datasets: 
the geodesic and normalization components consistently contribute the most to accuracy gains, 
confirming the universality of the proposed manifold-aware design.

\section{Related Work}

\subsection{Text-Attributed Graph Neural Networks}

Text-attributed graph neural networks (TAGNNs) aim to integrate textual semantics and graph topology into unified representations.
Early methods such as TextGCN~\cite{yao2019textgcn} model word/document relations as graphs and propagate textual signals via GCN-based architectures. 
Later studies including HeteGNN~\cite{wang2019heterogeneous} combine pretrained language models (PLMs) with graph aggregation, enabling richer semantic features. 
The CS-TAG benchmark~\cite{yan2023cstag} provides a systematic evaluation of TAG models across diverse domains and has become a standard testbed for recent research.
Recent progress further explores self-supervised and contrastive pretraining to better align textual and structural information.
For instance, ConGraT~\cite{park2023congrat} employs graph–text contrastive learning, GRENADE~\cite{yang2023grenade} introduces a graph-centric language model for self-supervised learning, and GAugLLM~\cite{zhou2024gaugllm} leverages large language models to generate semantic augmentations for contrastive objectives. 
NRUP~\cite{liu2023nrup} and RoSE~\cite{shi2024rose} introduce hierarchical and relation-sensitive architectures to strengthen text–structure coupling. 
BiGTex~\cite{xu2025bigtex} further designs a bi-directional fusion mechanism between text and graph signals.
Despite these advances, most TAGNNs still perform \emph{linear neighborhood aggregation} in Euclidean space, implicitly assuming that PLM embeddings reside in locally flat manifolds. 
However, contextualized representations are known to form curved and anisotropic semantic manifolds~\cite{ethayarajh2019contextual, timkey2021all}.
Linear aggregation over such spaces inevitably distorts geometry, leading to \textbf{semantic drift}—aggregated node embeddings deviating from the intrinsic manifold and losing semantic fidelity.
Prior studies on oversmoothing~\cite{rong2020dropedge, li2021training} partially relate to this issue but lack a geometric interpretation.
Our work provides the first quantitative framework to measure semantic drift in TAGs and introduces a manifold-consistent aggregation mechanism that explicitly mitigates it.

\subsection{Geometric-Aware Representation Learning}

Geometric deep learning~\cite{bronstein2021geometric,yang2023graph} has demonstrated the benefits of modeling data on non-Euclidean manifolds.
Hyperbolic and spherical graph neural networks~\cite{chami2019hgcn, bose2020latent, zhang2021lorentz, nickel2017poincare} map nodes into curved spaces to capture hierarchical or relational structures.
More recent developments include mixed-geometry and Riemannian GNNs~\cite{ganea2018hyperbolic, mathieu2020riemannian}, which perform message passing via log–exp maps or tangent-space aggregation.
Other geometric approaches~\cite{peng2021hyperbolic, li2025manifold} further explore manifold embeddings to improve representation quality.
These methods successfully exploit \emph{structural} geometry but generally ignore the \emph{semantic} geometry induced by pretrained text encoders.

In parallel, the geometry of PLM embeddings has attracted increasing attention. 
Empirical studies show that contextual embeddings occupy curved and anisotropic manifolds~\cite{ethayarajh2019contextual, timkey2021all}, suggesting that Euclidean operations such as mean pooling or linear interpolation distort intrinsic relationships between meanings.
Our work bridges this gap by introducing \textbf{GeoGNN}, a geodesic aggregation mechanism that performs message passing directly on the PLM-induced semantic manifold. 
By aligning graph propagation with the intrinsic curvature of textual representations through spherical normalization and log–exp mappings, GeoGNN preserves semantic geometry and effectively mitigates semantic drift across layers.

\section{Conclusion}
This paper investigates text-attributed graph learning from a geometric perspective. We reveal that conventional graph neural networks suffer from \textbf{semantic drift} when performing linear aggregation on the curved semantic manifolds induced by pretrained language models. To address this issue, we propose \textbf{Geodesic Aggregation}, a manifold-consistent message passing mechanism that operates along geodesics via log–exp mappings and spherical normalization. Built upon this principle, the resulting \textbf{GeoGNN} effectively mitigates semantic drift and yields consistent improvements across various datasets, text encoders, and downstream tasks. Beyond empirical gains, our study provides conceptual insights into how semantic geometry governs information propagation on Web-scale graphs. The proposed drift metric offers a quantitative tool to diagnose representation distortion, while the geodesic aggregation framework establishes a foundation for future research on geometry-aware text–graph reasoning and adaptive curvature modeling for heterogeneous Web data.


% \begin{acks}
% \end{acks}

\bibliographystyle{ACM-Reference-Format}
\bibliography{sample-base}

\newpage
\appendix

\section{Detailed Experimental Setup}
\label{appendix:setup}

\noindent\textbf{Datasets.}
We evaluate our model on four representative text-attributed graph datasets from the CS-TAG benchmark~\cite{yan2023cstag}: \textbf{Photo}, \textbf{Children}, \textbf{Arxiv}, and \textbf{History}. 
These datasets span diverse domains including e-commerce and academic citation networks, and all are formulated as both node classification and link prediction tasks. 
Each node is associated with a textual description, and edges represent co-view, co-purchase, or citation relationships. 
Dataset statistics are summarized in Table~\ref{tab:dataset_stats}.

\begin{table}[H]
\centering
\caption{Statistics of the four text-attributed graph datasets used in our experiments. 
The number of edges is reported before adding self-loops.}
\begin{tabular}{lccc}
\toprule
\textbf{Dataset} & \textbf{Nodes (Train/Val/Test)} & \textbf{Edges} & \textbf{Domain} \\
\midrule
Photo & 18,722 / 7,419 / 22,221 & 1,001,878 & E-Commerce \\
Children & 46,125 / 15,375 / 15,375 & 3,109,156 & E-Commerce \\
Arxiv & 90,941 / 29,799 / 48,603 & 2,332,486 & Academic \\
History & 24,930 / 8,310 / 8,311 & 717,148 & E-Commerce \\
\bottomrule
\end{tabular}
\label{tab:dataset_stats}
\end{table}

\noindent\textbf{Text Encoder.}
For node-level text representation, we employ a diverse set of pretrained language models (PLMs) covering different model families and scales, including 
\textit{RoBERTa-base}, 
\textit{MiniLM-L6-v2}, 
\textit{DistilBERT-base-uncased}, 
\textit{BERT-base-uncased}, 
\textit{RoBERTa-large}, 
\textit{BERT-large-uncased}, 
and \textit{Sentence-T5-large}.
All PLMs are kept \emph{frozen} during training to ensure that performance differences arise solely from the graph aggregation mechanisms rather than encoder finetuning. 
For each node, we obtain a fixed textual embedding by mean-pooling the last-layer token embeddings of the input text. 
These frozen PLM embeddings serve as the initial node features for all graph models in our experiments.

\medskip
\noindent\textbf{Baselines.}
We compare our proposed \textbf{GeoGNN} with nine representative graph neural networks:
GCN~\cite{kipf2017semi}, GAT~\cite{velickovic2018graph}, GraphSAGE~\cite{hamilton2017inductive}, 
GIN~\cite{xu2018powerful}, SGC~\cite{wu2019simplifying}, RevGAT~\cite{li2021training}, 
APPNP~\cite{gasteiger2018predict}, JKNet~\cite{xu2018representation}, 
and a non-graph \textbf{MLP} baseline. 
All models share the same frozen PLM-based features to isolate the geometric effect of different aggregation mechanisms. 
GeoGNN differs from these methods by replacing Euclidean message passing with geodesic aggregation that preserves the semantic manifold of PLM embeddings.

\medskip
\noindent\textbf{Training Configuration.}
All experiments are implemented in PyTorch and DGL.
We train each model for 1000 epochs using the Adam optimizer with a learning rate of $1\mathrm{e}{-3}$, dropout rate of $0.5$, and batch normalization disabled. 
Each experiment is repeated five times with different random seeds, and we report the average performance.
For GeoGNN, the attention temperature $\tau$ and geodesic step size $\alpha$ are treated as fixed hyperparameters and selected via grid search from 
$\{0.0001, 0.001, 0.01, 0.1, 1, 10, 100\}$. 
All experiments are conducted on NVIDIA H200 GPUs.

\medskip
\noindent\textbf{Evaluation.}
We follow the official CS-TAG data splits and use node classification accuracy as the main evaluation metric. 
For link prediction task, we split edges as $60\%/20\%/20\%$ for train/val/test. 
All models are trained and evaluated under identical protocols for a fair comparison across encoders and aggregators.

\section{Algorithm for Measuring Semantic Drift}
\label{appendix:drift_algorithm}

This section details the procedure for quantitatively measuring \textbf{semantic drift} described in Section~\ref{sec:drift}.
The algorithm computes node-level and graph-level drift scores using a local PCA-based approximation of the semantic manifold.
It reconstructs each GNN-updated embedding from the tangent subspace estimated in the pretrained language model (PLM) space,
and measures the normalized reconstruction error as the drift magnitude.
The complete algorithm is summarized in Algorithm~\ref{alg:semantic_drift}.

\begin{algorithm}[t]
\caption{Local PCA-based Measurement of Semantic Drift (Eqs.~\ref{eq:local_pca_error}--\ref{eq:normalized_drift})}
\label{alg:semantic_drift}
\begin{algorithmic}[1]
\Require 
PLM embeddings $\mathbf{X}_{\text{PLM}} = [\mathbf{x}_1,\dots,\mathbf{x}_n]$, \\
GNN embeddings $\mathbf{H}^{(l+1)} = [\mathbf{h}_1^{(l+1)}, \dots, \mathbf{h}_n^{(l+1)}]$, \\
number of neighbors $k$, PCA rank $r$.
\Ensure 
Node-wise drift scores $\{ D_i \}_{i=1}^{n}$ and mean drift $\bar{D}$.
\vspace{3pt}

\For{each node $v_i \in \mathcal{V}$}
    \State Find top-$k$ nearest neighbors 
    $\mathcal{N}_i = \{\mathbf{x}_j \mid j \in \text{top-}k(i)\}$ 
    in PLM feature space (cosine distance).
    \State Compute local mean $\bar{\mathbf{x}}_i$ and mean-centered data
    $\mathbf{Z}_i = \mathcal{N}_i - \bar{\mathbf{x}}_i$.
    \State Fit rank-$r$ PCA to $\mathbf{Z}_i$ to obtain local tangent subspace $\mathcal{T}_i$.
    \State Center GNN embedding $\mathbf{z}_i = \mathbf{h}_i^{(l+1)} - \bar{\mathbf{x}}_i$.
    \State Project and reconstruct using PCA inverse transform (as in Eq.~\ref{eq:local_pca_error}).
    \State Compute reconstruction error $E_i$ following Eq.~\ref{eq:local_pca_error}.
    \State Normalize to obtain drift score $D_i$ according to Eq.~\ref{eq:normalized_drift}.
\EndFor
\State Compute global mean drift $\bar{D} = \frac{1}{|\mathcal{V}|}\sum_i D_i$.
\State \Return $\{D_i\}_{i=1}^{n}$, $\bar{D}$.
\end{algorithmic}
\end{algorithm}

\end{document}